\documentclass[letterpaper]{article} 
\usepackage{aaai2026}  
\usepackage{times}  
\usepackage{helvet}  
\usepackage{courier}  
\usepackage[hyphens]{url}  
\usepackage{graphicx} 
\usepackage{natbib}  
\usepackage{caption} 
\usepackage{algorithm}
\usepackage{algorithmic}

\usepackage{multirow}
\usepackage{booktabs} 
\usepackage{makecell}

\usepackage{newfloat}
\usepackage{listings}
\DeclareCaptionStyle{ruled}{labelfont=normalfont,labelsep=colon,strut=off} 
\floatstyle{ruled}
\newfloat{listing}{tb}{lst}{}
\floatname{listing}{Listing}
 \nocopyright 

\title{GraphIF: Enhancing Multi-Turn Instruction Following for 
 Large Language Models with Relation Graph Prompt}
\author{
    Zhenhe Li\textsuperscript{\rm 1},
    Can Lin\textsuperscript{\rm 1},
    Ling Zheng\textsuperscript{\rm 1},
    Wen-Da Wei\textsuperscript{\rm 2},
    Junli Liang\textsuperscript{\rm 1},
    Qi Song\textsuperscript{\rm 1}\thanks{Corresponding author}
}
\affiliations{
    \textsuperscript{\rm 1}University of Science and Technology of China\\
    \textsuperscript{\rm 2}School of Artificial Intelligence, Nanjing University, China\\
    \{lizh2002, can.lin, lingzheng, jlliang\}@mail.ustc.edu.cn, weiwd@lamda.nju.edu.cn,  qisong09@ustc.edu.cn

}

\usepackage{bibentry}

\begin{document}

\maketitle

\begin{abstract}
Multi-turn instruction following is essential for building intelligent conversational systems that can consistently adhere to instructions across dialogue turns. However, existing approaches to enhancing multi-turn instruction following primarily rely on collecting or generating large-scale multi-turn dialogue datasets to fine-tune large language models (LLMs), which treat each response generation as an isolated task and fail to explicitly incorporate multi-turn instruction following into the optimization objectives. As a result, instruction-tuned LLMs often struggle with complex long-distance constraints. In multi-turn dialogues, relational constraints across turns can be naturally modeled as labeled directed edges, making graph structures particularly suitable for modeling multi-turn instruction following. Despite this potential, leveraging graph structures to enhance the multi-turn instruction following capabilities of LLMs remains unexplored. To bridge this gap, we propose GraphIF, a plug-and-play framework that models multi-turn dialogues as directed relation graphs and leverages graph prompts to enhance the instruction following capabilities of LLMs. GraphIF comprises three key components: (1) an agent-based relation extraction module that captures inter-turn semantic relations via action-triggered mechanisms to construct structured graphs; (2) a relation graph prompt generation module that converts structured graph information into natural language prompts; and (3) a response rewriting module that refines initial LLM outputs using the generated graph prompts. Extensive experiments on two long multi-turn dialogue datasets demonstrate that GraphIF can be seamlessly integrated into instruction-tuned LLMs and leads to significant improvements across all four multi-turn instruction-following evaluation metrics.
\end{abstract}

\begin{links}
    \link{Code}{https://github.com/sstillzh/GraphIF}
\end{links}

\begin{figure}
    \centering
    \includegraphics[width=1.0\linewidth]{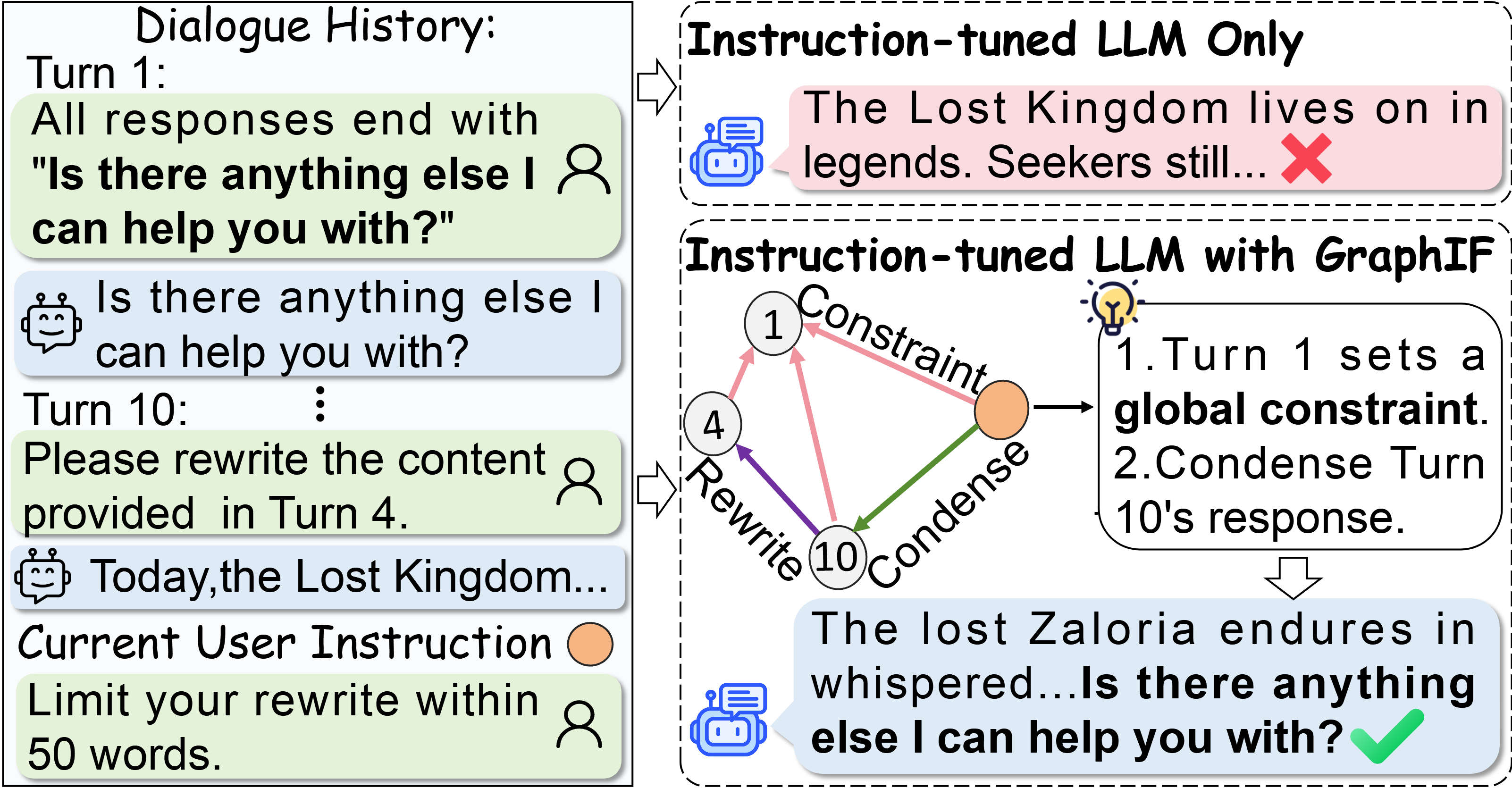}
    \caption{
    A comparison of two types of methods:
Instruction-tuned LLM only and our
proposed GraphIF that uses graph structure to enhance the multi-turn instruction following.
    }
    \label{fig:introduction}
\end{figure}

\section{Introduction}
Large Language Models (LLMs) \cite{achiam2023gpt,dubey2024llama,team2024qwen2} have demonstrated exceptional performance in dialogue systems. As conversational AI becomes increasingly important, the ability to understand and follow user instructions is crucial for effective interaction \cite{chiang2023vicuna,zheng2023judging}. Real-world conversations pose a challenge as users typically distribute requirements across multiple dialogue turns, making multi-turn instruction following essential for maintaining dialogue coherence and consistency.

Recent efforts to enhance multi-turn instruction following have primarily focused on collecting \cite{chiang2023vicuna,zhao2024wildchat} or generating \cite{cao2025condor,gao2025strategic} multi-turn dialogue datasets to fine-tune LLMs. However, even well-tuned LLMs struggle with complex long-distance constraints \cite{kwan2024mt}. As illustrated in Figure \ref{fig:introduction}, LLMs often forget instructions established in earlier turns, such as ending responses with specific phrases. This limitation stems from the mismatch between autoregressive training and the non-sequential dependencies in multi-turn dialogues—current instruction-tuning methods treat each response generation as an isolated task without explicitly modeling cross-turn constraints \cite{zhang2025survey}.


While incremental training could mitigate this issue, it incurs substantial computational costs and suffers from weak generalization due to data limitations \cite{li2025beyond}. This raises a fundamental question: can we develop training-free approaches that enable LLMs to satisfy inter-turn relational constraints without parameter updates?

In order to satisfy inter-turn relational constraints, the LLM should have the ability to remember the previous dialogues, correctly identify the related contexts, and then accordingly generate answers to the instruction in the current turn. This imposes two challenges: 1) \textbf{How to extract the semantic relations between different turns?} As illustrated in Figure \ref{fig:introduction}, turn 1 introduces a global constraint requiring all responses to end with a specific sentence, and the current instruction requests compression of turn 10's response. Extracting such inter-turn relations remains non-trivial. 2) \textbf{How to leverage the extracted relations to generate correct responses?} Once the model identifies the specific relations linking the current instruction to turns 1 and 10, how can these captured relations be effectively utilized to generate responses that satisfy all relevant constraints?

Graph structure has been proven to be able to represent the semantic relations in scenarios including  multi-document QA \cite{wang2024knowledge,edge2024local} and long-context QA \cite{li2024graphreader}. In multi-turn dialogues, the relational constraints between different dialogue turns can be naturally represented through labeled directed edges, making graph structures particularly suitable for modeling multi-turn instruction following \cite{li2025structflowbench}. Despite existing work on using graph structures to generate multi-turn dialogue datasets \cite{li2025structflowbench}, leveraging graph structures to enhance multi-turn instruction following remains unexplored. 

To address the aforementioned challenges and bridge the gap, we propose GraphIF, a training-free and plug-and-play framework that models multi-turn dialogues as directed relation graphs and leverages graph prompts to enhance multi-turn instruction following for LLMs. We use graph structures to uniformly model inter-turn relations as structured information. As shown in Figure \ref{fig:introduction}, each node in the relation graph represents a dialogue turn, and labeled directed edges capture constraint relations between connected turns. The directed edge \texttt{<Current Turn, Global Constraint Imposed By, Turn 1>} indicates that Turn 1 imposes a global constraint that the current response must satisfy. Key components of GraphIF are as follows:

(1) We design an agent-based relation extraction module. Directly extracting semantic relations between dialogue turns is difficult, especially for complex long-distance constraints \cite{bai2024mt,sun2024parrot}. Inspired by agent-based decomposition paradigms \cite{yao2023react,erdogan2025plan}, we design an agent-based relation extraction module that employs LLM to alternate between action identification and action execution phases to iteratively construct the dialogue relation graph.

(2) We design two modules to leverage the extracted relations for generating correct responses. We first design a relation graph prompt generation module that converts the structured graph information into natural language prompts, explicitly articulating inter-turn relations and their corresponding constraints. Finally, we design an initial response rewrite module that leverages the constraints articulated in the graph prompts to refine the initial responses generated by LLMs. This refinement ensures better adherence to multi-turn dialogue constraints.


Given that existing multi-turn instruction following datasets have limited dialogue turns and oversimplified inter-turn relations \cite{zheng2023judging,he2024multi,bai2024mt}, we construct two new datasets with extended dialogues and complex semantic relations based on two prior benchmarks \cite{kwan2024mt,li2025structflowbench}.
Comprehensive experiments on the two datasets demonstrate that current fine-tuned LLMs exhibit suboptimal performance when confronted with complex constraints. Our GraphIF framework can be seamlessly integrated into existing LLMs and achieves significant improvements across all four multi-turn instruction-following evaluation metrics.

We summarize our contributions as follows:
\begin{itemize}
    \item We propose GraphIF, a training-free and plug-and-play framework that explicitly models inter-turn relations through graph structures and generates graph prompts to refine initial LLM responses, thereby enhancing multi-turn instruction following capabilities of LLMs.
    \item We design an agent-based relation extraction module that iteratively performs action identification and execution to construct dialogue relation graphs, addressing the challenge of difficult relation extraction. To leverage the extracted relations for generating correct responses, we develop a relation graph prompt generation module that converts graph structures into natural language prompts, and a response rewrite module that uses graph prompts to refine LLM outputs.
    \item Extensive experiments on two long multi-turn dialogue datasets demonstrate that GraphIF can be seamlessly integrated into existing instruction-tuned LLMs and leads to significant improvements across all four multi-turn instruction-following evaluation metrics.
\end{itemize}

\section{Related Work}

\paragraph{Instruction Fine-tuning for LLMs.} 
Current approaches for enhancing multi-turn instruction following mainly use supervised fine-tuning with instruction datasets. Some methods curate real-world user-LLM interactions \cite{wang2023openchat,zhao2024wildchat}, while others generate synthetic dialogues \cite{ding2023enhancing,wu2025instruct}. Parrot \cite{sun2024parrot} addresses anaphora and ellipsis by training specialized models, and ConsistentChat \cite{chen2025consistentchat} tackles consistency through skeleton-guided generation. However, these instruction-tuning methods overlook explicit modeling of inter-turn relational structures.

\begin{figure*}[t]
\centering
\includegraphics[width=1.0\textwidth]{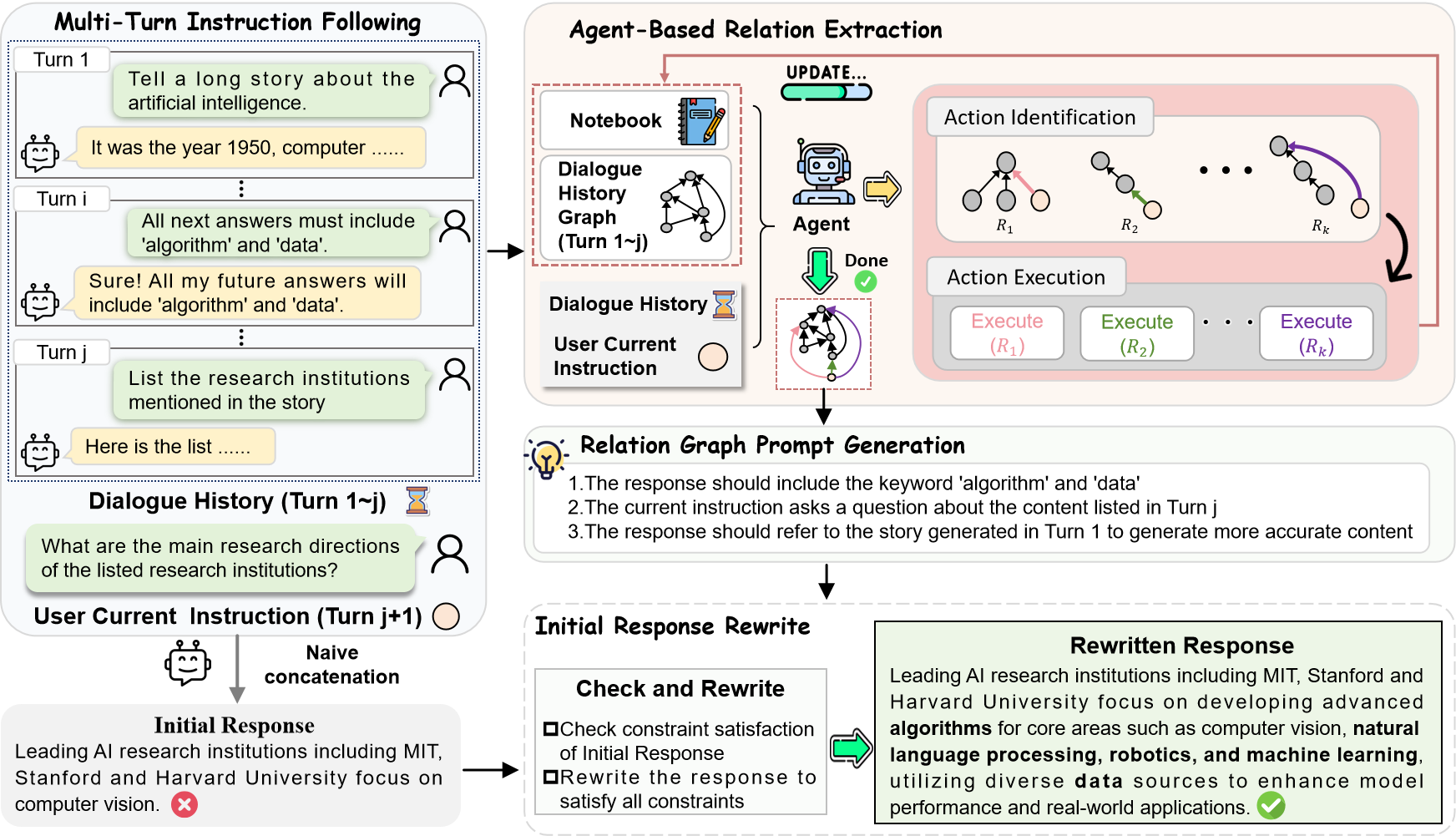} 
\caption{Framework overview of GraphIF. Given dialogue history and current user instruction, GraphIF first extracts semantic relations between dialogues through the \textit{Agent-Based Relation Extraction} module, then employs \textit{Relation Graph Prompt Generation} to generate constraint-aware prompts, and finally uses the \textit{Initial Response Rewrite} module to refine the initial response.}
\label{fig:method}
\end{figure*}

\paragraph{Graph-Augmented Generation with LLMs.} Recent research has increasingly explored the integration of graph structures to enhance the capabilities of LLMs \cite{jimenez2024hipporag,li2024graphreader,lin2025rje}. GraphRAG \cite{edge2024local} and LightRAG \cite{guo2024lightrag} construct cross-document knowledge graphs to enable knowledge-enhanced retrieval for response generation. KGP \cite{wang2024knowledge} addresses multi-document question-answering tasks by constructing passage-level directed graphs to aggregate relevant information. However, no existing research has applied graph structures to enhance the multi-turn instruction following for LLMs.

\section{Preliminary}
\textbf{Multi-Turn Instruction Following} We formally define the \textit{multi-turn instruction following task} as $\mathcal{D} = \{ \left( \mathcal{H}_t ,\mathcal{I}_t) \right\}_{t=1}^{M}$, where $M$ represents total number of dialogue turns ($M>1$), $\mathcal{I}_t$ represents the user instruction in $t$-th dialogue turn, and $\mathcal{H}_t=\left\{ \left<\mathcal{I}_k, \mathit{RES}_k \right> \right\}_{k=1}^{t-1}$ represents the dialogue history before turn $t$, where $\mathit{RES}_k$ means the response generated by LLM in $k$-th dialogue turn. Specifically, $\mathcal{H}_1$ is empty in the first dialogue turn, and the LLM directly generates a response to $\mathcal{I}_1$. For definitional rigor, our subsequent discussion focuses on scenarios where $t \geq 2$, ensuring that the dialogue history $\mathcal{H}_t$ is non-empty. Given an instruction $\mathcal{I}_t$ and dialogue history $\mathcal{H}_{t}$, the purpose of multi-turn instruction following task is to generate an overall response $RES_t$ that aligns with the given context while maintaining semantic consistency across turns.

\textbf{Relation Graph in GraphIF}  We employ the directed relation graph to model the multi-turn instruction-following dialogue scenario. In the graph, each vertex corresponds to a dialogue turn, while the directed edges between vertices capture the semantic relations across dialogue turns. Formally, let $\mathcal{G}=\{\mathcal{V},\mathcal{E},\mathcal{R}\}$ be the constructed relation graph, where each node $v_i \in \mathcal{V}$ represents a complete dialogue turn $<\mathcal{I}_i,\mathit{RES}_i>$. And $\mathcal{E} \subset \mathcal{V} \times \mathcal{R} \times \mathcal{V}$ represents the edge set, where $\mathcal{R}$ denotes the predefined relation labels between dialogue turns.

\section{Methodology}
In this section, we introduce the technical details of GraphIF. Figure \ref{fig:method} illustrates the overall framework. GraphIF mainly contains three components: (1) Agent-Based Relation Extraction module, which extracts relations between dialogue turns to construct the structured graph, (2) Relation Graph Prompt Generation module, which transforms the structured graph information into natural language prompts, (3) Initial Response Rewrite module, which leverages the graph prompt to refine the initial response generated by LLMs.
\subsection{Agent-Based Relation Extraction}
In multi-turn instruction following scenarios, complex contextual semantic relations exist between current user instructions and dialogue history. To model these interactions, we explicitly identify and extract semantic relations between current instructions and historical dialogue turns, constructing a semantic relation graph. Formally, given user instruction $\mathcal{I}_{t}$ and dialogue history $\mathcal{H}_{t}$ at turn $t$, we identify specific semantic relations between $\mathcal{I}_{t}$ and targeted turns in $\mathcal{H}_{t}$.

Given complex relational dependencies between current instructions and multiple dialogue turns, single-step relation extraction is insufficient. Inspired by agent-based problem decomposition paradigms \cite{yao2023react,erdogan2025plan}, we propose an iterative agent-based module for progressive relation extraction. Since different dialogue relations require tailored identification approaches, our module decomposes relation extraction into two complementary subtasks: \textbf{Action Identification} and \textbf{Action Execution}.

In Action Identification, we introduce an action-triggered mechanism to identify corresponding relations. Formally, given the predefined dialogue relation set $\mathcal{R}=\{r_1,r_2, \dots,r_m \}$, each relation label $r_i \in \mathcal{R}$  corresponds to an action $a_i \in \mathcal{A}=\{a_1,a_2,\dots,a_m\}$ that uniquely identifies it.  During the iterative relation extraction process, $a_{ts'}$ represents the action identified at timestep $s'$ of turn $t$, indicating the identification of the corresponding relation $r_{ts'}$. The iterative process continues until the termination action Done is identified, indicating that all relevant relations have been extracted.

In Action Execution, each action $a_i$ maintains a dedicated implementation function $Execute(a_i)$ to extract dialogue turns satisfying the identified relations with the current instruction. For the identified action $a_{ts'}$, we typically denote the execution result as $E_{ts'}$, which represents the set of historical dialogue turns that satisfy the relation $r_{ts'}$ with the current instruction. These identified turns define directed edges in the relation graph, where each edge points from the current instruction to a historical dialogue turn, representing the relation $r_{ts'}$ between them.

Additionally, to enable incremental updates while avoiding redundant extraction, we introduce a dynamic notebook mechanism $\mathcal{N}=\{a_{ts'} \rightarrow E_{ts'} \}$ that maintains relation mappings across iterations, allowing the agent to progressively construct the dialogue relation graph.

In the following, we detail the concrete implementation of the action identification and action execution phases. 
\subsubsection{Action Identification}
Given the potential existence of multiple relations between $\mathcal{I}_t$ and $\mathcal{H}_{t}$, the agent employs an iterative updating mechanism, where the action identification module identifies the most critical relation at each timestep and updates the notebook accordingly. Specifically, at timestep $s$, based on the extracted information maintained in the notebook $\mathcal{N}$, we prompt the LLM to identify the next relation instance from the dialogue history and return the corresponding action $a_{ts}$ (return Done action if all required content has been extracted):
\begin{equation}
    a_{ts} = LLM(\mathcal{I}_t,\mathcal{H}_{t},\mathcal{N}).
\end{equation}

\subsubsection{Action Execution} Upon action identification, the identified action $a_{ts}$ corresponds to relation $r_{ts} \in \mathcal{R}$, and the corresponding implementation function $Execute(a_{ts})$ will be executed. Analogous to the action identification phase, we leverage the notebook $\mathcal{N}$ to maintain the memory of extracted relations, thereby eliminating redundant relation recognition. Typically, we prompt LLM to locate the $E_{ts}$:
\begin{equation}
    E_{ts} = LLM(\mathcal{I}_t,\mathcal{H}_{t},r_{ts},\mathcal{N}).
\end{equation}

After finishing locating $E_{ts}$, we update the notebook $\mathcal{N}$ and dialogue relation graph $\mathcal{G}$:
\begin{equation}
    \mathcal{N} = \mathcal{N} \cup (a_{ts} \rightarrow E_{ts}),
\end{equation}
\begin{equation}
    \mathcal{G} = \mathcal{G} + \bigcup\limits_{i \in E_{ts}} (v_t,r_{ts},v_i).
\end{equation}
Here, we update the notebook with the identified action $a_{ts}$ and the relevant dialogue turns $E_{ts}$. Additionally, each dialogue turn is connected to the current user instruction by an edge of relation type $r_{ts}$.

\subsubsection{Implementation} We define a set of actions that characterize representative dialogue relations. Unlike previous benchmarks \cite{kwan2024mt,li2025structflowbench}, we define a more inclusive and coarse-grained set of relations that better capture the dialogue constraints in real-world scenarios.

\textit{Identify\_Global\_Constraint}: Identifies the current instruction as a global constraint for subsequent interactions, adding it to the global constraint set.

\textit{Identify\_Context\_Anchored}: Identifies specific historical dialogue content that the instruction semantically relies on or logically connects to. The implementation locates the most relevant unrecorded dialogue turn.

\textit{Identify\_Modify}: Identifies specific historical dialogue content that the current instruction refines or extends, built upon context-anchored relations with additional requirements. The implementation follows the same approach as context-anchored relation extraction.

\textit{Identify\_Summary}: Identifies the instruction that requests a summary of specific historical dialogue turns. The implementation locates all dialogue turns to be summarized.

\textit{New\_Topic}: Identifies instructions that introduce entirely new topics unconnected to the current dialogue context. The implementation determines whether the topic shifts back to a previous topic or initiates a new topic.

\textit{Done}: Indicates complete relation extraction termination.

\subsection{Relation Graph Prompt Generation}
We convert the extracted dialogue relations into interpretable prompt formats to enhance contextual understanding of LLMs.

Concretely, we construct the graph prompt $\mathcal{P}_g$ through a structured concatenation process. For each identified semantic relation $r_{ts}$, we systematically integrate three parts: (1) the formal definition of $r_{ts}$, (2) the specific description of how this relation connects to the current instruction $\mathcal{I}_t$, i.e., the specific constraints that the response should satisfy and (3) dialogue content of the corresponding turns.

Figure \ref{fig:method} illustrates the core content of $\mathcal{P}_g$, including the identified semantic relations and the constraints that the response should satisfy.

\begin{table*}[t!]
    \centering
    \small
    \begin{tabular}{c|c|cccc|cccc}
        \toprule
        \multirow{2}{*}{\textbf{Backbone Model}}  &  \multirow{2}{*}{\textbf{Method}} & \multicolumn{4}{c|}{\textbf{MT-Eval*(\%)}} & \multicolumn{4}{c}{\textbf{StructFlowBench*(\%)}}\\
        & &  \textbf{CSR} & \textbf{ISR} & \textbf{DRFR} & \textbf{WCSR} & \textbf{CSR} & \textbf{ISR} & \textbf{DRFR} & \textbf{WCSR} \\
        
        \midrule
        \multirow{4}{*}{Qwen2.5-7B-Instruct} & LLM-only  & 80.22 & 51.30 & 79.70 & 79.57 & 70.70 & 23.02 & 71.22 & 67.07 \\
        & +MemoryBank & 70.29 & 30.87 & 69.84 & 67.88 & 74.63 & 29.99 & 74.92 & 70.58 \\
        & +MemoChat & 59.42 & 16.52 & 58.79 & 56.35 & 72.87 & 27.08 & 73.23 & 68.72 \\
        & +GraphIF & \textbf{91.30} & \textbf{76.96} & \textbf{91.06} & \textbf{90.72} & \textbf{89.46} & \textbf{69.25} & \textbf{89.47} & \textbf{88.60} \\

        \midrule
        \multirow{4}{*}{Llama-3.1-8B-Instruct} & LLM-only  & 67.03 & 27.51 & 67.88 & 64.22 & 71.03 & 29.32 & 72.75 & 67.98 \\
        & +MemoryBank & 67.17 & 31.74 & 66.97 & 67.74 & 74.37 & 35.85 & 75.48 & 70.83 \\
        & +MemoChat & 56.30 & 15.22 & 55.45 & 53.97 & 73.51 & 32.07 & 74.29 & 69.86 \\
        & +GraphIF & \textbf{91.27} & \textbf{80.35} & \textbf{90.87} & \textbf{92.58} & \textbf{88.91} & \textbf{70.29} & \textbf{89.38} & \textbf{88.04} \\

        \midrule
        \multirow{4}{*}{Hermes-3-Llama-3.1-8B} & LLM-only  & 76.74 & 48.70 & 76.21 & 76.09 & 68.39 & 23.55 & 69.74 & 65.11 \\
        & +MemoryBank & 68.41 & 34.35 & 68.33 & 69.07 & 69.20 & 23.57 & 70.03 & 64.98 \\
        & +MemoChat & 54.93 & 11.30 & 53.94 & 52.26 & 68.79 & 24.64 & 69.89 & 64.76 \\
        & +GraphIF & \textbf{86.96} & \textbf{70.43} & \textbf{86.52} & \textbf{86.64} & \textbf{78.74} & \textbf{46.74} & \textbf{79.73} & \textbf{76.54} \\

        \midrule
        \multirow{4}{*}{Llama-3.1-Storm-8B} & LLM-only  & 80.36 & 56.96 & 81.36 & 80.87 & 74.07 & 34.12 & 74.99 & 70.75 \\
        & +MemoryBank & 66.16 & 28.70 & 65.45 & 66.35 & 73.92 & 31.94 & 74.37 & 70.02 \\
        & +MemoChat & 55.36 & 13.04 & 54.39 & 53.59 & 71.55 & 30.60 & 71.62 & 67.78 \\
        & +GraphIF & \textbf{93.62} & \textbf{81.30} & \textbf{93.33} & \textbf{94.87} & \textbf{86.92} & \textbf{64.18} & \textbf{86.83 } & \textbf{85.32} \\

        \midrule
        \multirow{4}{*}{Llama-3.1-Tulu-3.1-8B} & LLM-only  & 74.13 & 44.35 & 75.30 & 72.70 & 80.92 & 47.85 & 81.93 & 78.78 \\
        & +MemoryBank & 73.70 & 43.04 & 73.03 & 73.45 & 78.51 & 42.24 & 78.88 & 75.32 \\
        & +MemoChat & 50.65 & 19.57 & 51.36 & 49.83 & 64.71 & 25.13 & 66.46 & 61.85 \\
        & +GraphIF & \textbf{88.48} & \textbf{76.96} & \textbf{88.03} & \textbf{89.68} & \textbf{88.68} & \textbf{69.71} & \textbf{88.98} & \textbf{87.90} \\
        \bottomrule
    \end{tabular}
    \caption{Performance comparison of different methods across five backbone LLMs on MT-Eval* and StructFlowBench* datasets. LLM-only: direct concatenation of dialogue history and user instruction for response generation; Others: integration of respective methods into LLM. Best results are highlighted in bold.}
    \label{tab:main_results}
\end{table*}

\begin{table}[t]
    \centering
    \small
    \setlength{\tabcolsep}{1mm}
    \begin{tabular}{cccccc}
        \toprule
        \multirow{2}{*}{\textbf{Model}} & \multirow{2}{*}{\textbf{Method}} & \multicolumn{4}{c}{\textbf{MT-Eval*(\%)}} \\
        & &  \textbf{CSR} & \textbf{ISR} & \textbf{DRFR} & \textbf{WCSR}  \\
        \midrule
        \multirow{2}{*}{\makecell[c]{\textbf{Qwen2.5-3B}\\\textbf{-Instruct}}}
        & LLM-Only & 59.42 & 9.56 & 59.09 & 56.99\\
        & +GraphIF & \textbf{75.43} & \textbf{46.09} & \textbf{75.45} & \textbf{75.51}\\
        \midrule
        \multirow{2}{*}{\makecell[c]{\textbf{Qwen2.5-7B}\\\textbf{-Instruct}}}
        & LLM-Only & 80.22 & 51.30 & 79.70 & 79.57\\
        & +GraphIF & \textbf{91.30} & \textbf{76.96} & \textbf{91.06} & \textbf{90.72}\\
        \midrule
        \multirow{2}{*}{\makecell[c]{\textbf{Qwen2.5-14B}\\\textbf{-Instruct}}}
        & LLM-Only & 90.22 & 75.65 & 90.00 & 92.61\\
        & +GraphIF & \textbf{93.62} & \textbf{82.17} & \textbf{93.48} & \textbf{95.71}\\
        \bottomrule
    \end{tabular}
    \caption{Performance comparison of Qwen2.5 models of different scales as backbone models on the MT-Eval*.}
    \label{tab:mteval}
\end{table}
\subsection{Initial Response Rewrite}
We generate the initial response by directly concatenating the dialogue history with the user instruction, which represents the implementation of LLM in real-world applications: 
\begin{equation}
    \mathit{RES}_{initial} = LLM([\mathcal{H}_{t},\mathcal{I}_t]).
\end{equation}
As illustrated in Figure \ref{fig:method}, $\mathit{RES}_{initial}$ fails to incorporate the two specified keywords and exhibits inaccuracies, omitting crucial information from the original story.

Then we leverage the constructed graph prompt to refine the initial response:
\begin{equation}
    \mathit{RES}_{t} =  \mathrm{Rewrite}(RES_{initial},\mathcal{P}_g).
\end{equation}
As illustrated in Figure \ref{fig:method}, the LLM identifies unsatisfied constraints in the $\mathit{RES}_{initial}$ by examining the content of $\mathcal{P}_g$, and subsequently performs targeted rewriting to generate a refined response $\mathit{RES}_{t}$ that incorporates the two required keywords and addresses the key information missing from the $\mathit{RES}_{initial}$.

\begin{table*}[t]
    \centering
    \small
    \begin{tabular}{l|l|l|cccc}
        \toprule
        \multirow{2}{*}{\textbf{Dataset}} & \multirow{2}{*}{\textbf{Model}} & \multirow{2}{*}{\textbf{Method}} & \multicolumn{4}{c}{\textbf{Results(\%)}} \\
        
        & & & \textbf{CSR} & \textbf{ISR} & \textbf{DRFR} & \textbf{WCSR}  \\
        \midrule
        \multirow{6}{*}{\textbf{MT-Eval*}}
        & \multirow{3}{*}{\textbf{Qwen2.5-7B-Instruct}}
        & GraphIF & \textbf{91.30} & \textbf{76.96} & \textbf{91.06} & \textbf{90.72} \\
        & &  \ \ w/o Relation Extraction Agent & 67.17 & 21.74 & 66.52 & 63.48\\
        & &  \ \ w/o Graph Prompt & 85.80 & 63.48 & 85.45 & 84.14 \\
        \cmidrule(lr){2-7}
        & \multirow{3}{*}{\textbf{Llama-3.1-8B-Instruct}}
        & GraphIF & \textbf{91.27} & \textbf{80.35} & \textbf{90.87} & \textbf{92.58} \\
        & &  \ \ w/o Relation Extraction Agent & 73.77 & 40.87 & 73.33 & 72.72 \\
        & &  \ \ w/o Graph Prompt & 84.93 & 63.32 & 84.47 & 85.07 \\
        \midrule
        \multirow{6}{*}{\textbf{StructFlowBench*}}
        & \multirow{3}{*}{\textbf{Qwen2.5-7B-Instruct}}
        & GraphIF & \textbf{89.46} & \textbf{69.25} & \textbf{89.47} & \textbf{88.60} \\
        & &  \ \ w/o Relation Extraction Agent & 73.36 & 27.71 & 74.11 & 69.53\\
        & &  \ \ w/o Graph Prompt & 75.93 & 30.95 & 75.74 & 72.34 \\
        \cmidrule(lr){2-7}
        & \multirow{3}{*}{\textbf{Llama-3.1-8B-Instruct}}
        & GraphIF & \textbf{88.91} & \textbf{70.29} & \textbf{89.38} & \textbf{88.04} \\
        & &  \ \ w/o Relation Extraction Agent & 81.74 & 55.08 & 81.99 & 79.61 \\
        & &  \ \  w/o Graph Prompt & 77.56 & 44.11 & 77.84 & 74.82 \\
        \bottomrule
    \end{tabular}
    \caption{The results of ablation study. ``w/o Relation Extraction Agent" refers to removing the agent module and directly extracting all inter-turn relations via one-time LLM inference, ``w/o Graph Prompt" denotes using only relevant dialogue content without semantic relation explanations.}
    \label{tab:ablation}
\end{table*}

\section{Experiments}

\subsection{Experimental Settings}
\subsubsection{Dataset} 
Existing multi-turn instruction-following datasets are limited by short dialogue turns (typically under eight turns) \cite{bai2024mt,he2024multi}, inadequately representing complex multi-turn instruction following scenarios. To address this, we construct evaluation datasets following two benchmarks that model inter-turn relations.
We design MT-Eval* by merging and manually verifying multiple instances from MT-Eval \cite{kwan2024mt}. Similarly, we leverage the customizable structural framework of StructFlowBench \cite{li2025structflowbench} to design StructFlowBench*. The two datasets we construct feature dialogues with over 20 turns, each containing multiple inter-turn relations within individual dialogue instances.

\subsubsection{Backbone Model}
We evaluate five instruction-tuned LLMs as backbone architectures, comprising two official models: Llama-3.1-8B-Instruct \cite{dubey2024llama} and Qwen2.5-7B-Instruct \cite{team2024qwen2}, along with three models that are fine-tuned and aligned through preference-based methods based on Llama-3.1-8B: Hermes-3-Llama-3.1-8B\footnote{\url{https://huggingface.co/NousResearch/Hermes-3-Llama-3.1-8B}}, Llama-3.1-Storm-8B\footnote{\url{https://huggingface.co/akjindal53244/Llama-3.1-Storm-8B}}, and Llama-3.1-Tulu-3.1-8B\footnote{\url{https://huggingface.co/allenai/Llama-3.1-Tulu-3.1-8B}} \cite{lambert2024tulu}.

\subsubsection{Baseline}
Due to the lack of robust frameworks specifically designed for multi-turn instruction-following enhancement, we select two representative memory-enhanced frameworks as baselines:

\begin{itemize}
    \item MemoryBank \cite{zhong2024memorybank} summarizes dialogue content as events and employs RAG mechanisms to retrieve relevant conversations, enhancing models' memory of key information. 
    \item MemoChat \cite{lu2023memochat} creates topic-indexed storage structures and summarizes related dialogues, using LLM-based retrieval to enhance response generation.
\end{itemize}

\subsubsection{Evaluation}
We adopt four instruction-following metrics to evaluate models from different perspectives: Constraint Satisfaction Rate (CSR), Instruction Satisfaction Rate (ISR) \cite{zhang2024cfbench}, Decomposed Requirements Following Ratio (DRFR) \cite{qin2024infobench}, and Weighted Constraint Satisfaction Rate (WCSR) \cite{li2025structflowbench}. Following recent benchmarks \cite{zheng2023judging,li2025structflowbench}, we utilize GPT-4o as the LLM judge with human verification. 


\subsubsection{Implementation Details}

We set the temperature parameter of the LLMs to 0.7, $top\_p$ to 0.8, and $top\_k$ to 20 for achieving a balance between response diversity and generation stability. For evaluation, we conduct three independent experimental runs and compute the average of the evaluation metrics as the results. All experiments are conducted on 4 A800-80GB GPUs.

\begin{figure}
    \centering
    \includegraphics[width=1.0\linewidth]{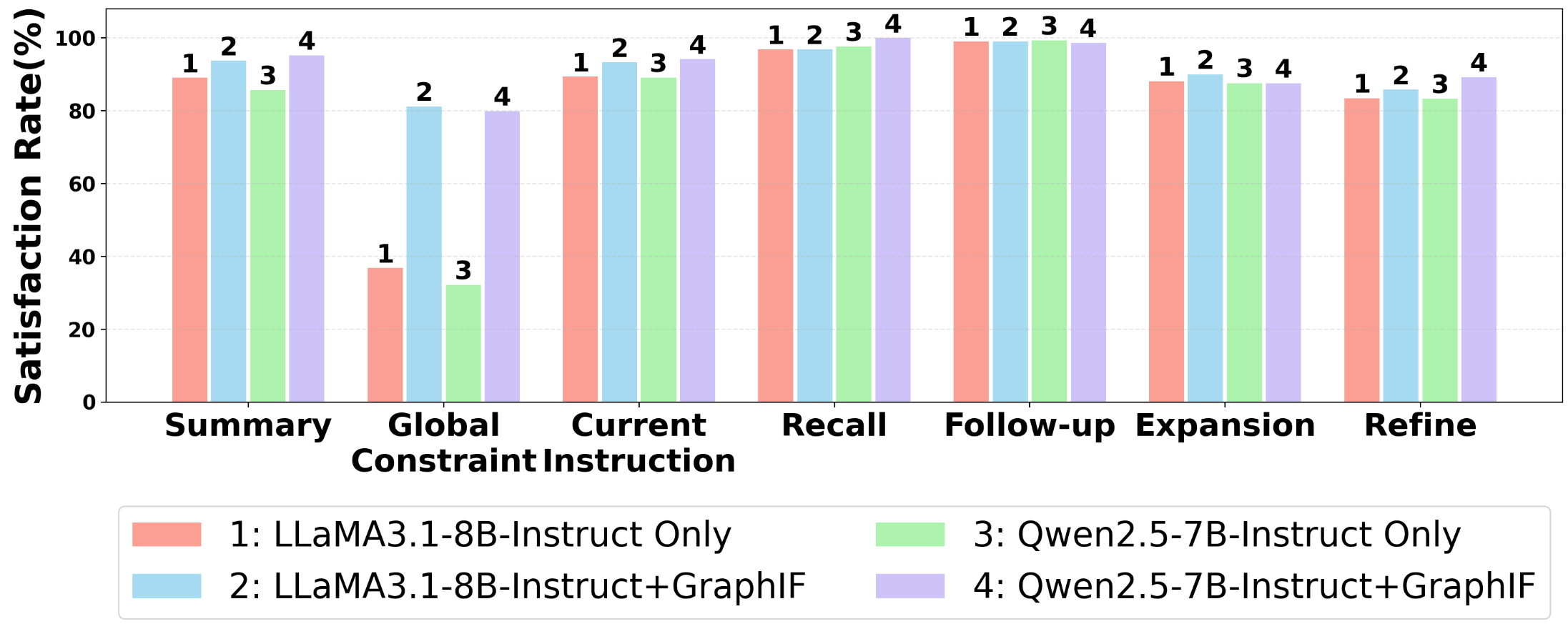}
    \caption{Detailed constraint satisfaction results across different constraint types on the StructFlowBench*.}
    \label{fig:constraint_structbench}
\end{figure}
\subsection{Main Results}
\subsubsection{Overall Performance Comparison.}
Table \ref{tab:main_results} presents the comprehensive evaluation of GraphIF against baseline methods on both MT-Eval* and StructFlowBench* datasets. Integrating GraphIF into instruction-tuned LLMs yields substantial improvements across all multi-turn instruction following metrics, with performance gains ranging from 7\% to 52\%. The most striking improvement appears in the Instruction Satisfaction Rate (ISR), the most stringent metric that scores only when responses satisfy all constraints within a dialogue turn. GraphIF achieves over 20\% improvement in ISR, indicating a significant increase in the proportion of dialogues that perfectly adhere to all specified constraints.

In contrast, memory-enhanced approaches (MemoryBank and MemoChat) fail to improve over vanilla instruction-tuned LLMs, showing negligible gains or even hurting performance. This highlights the fundamental limitations of memory-based methods in capturing complex relational structures in multi-turn dialogues.

\begin{figure*}
    \centering
    \small
    \includegraphics[width=0.9\linewidth]{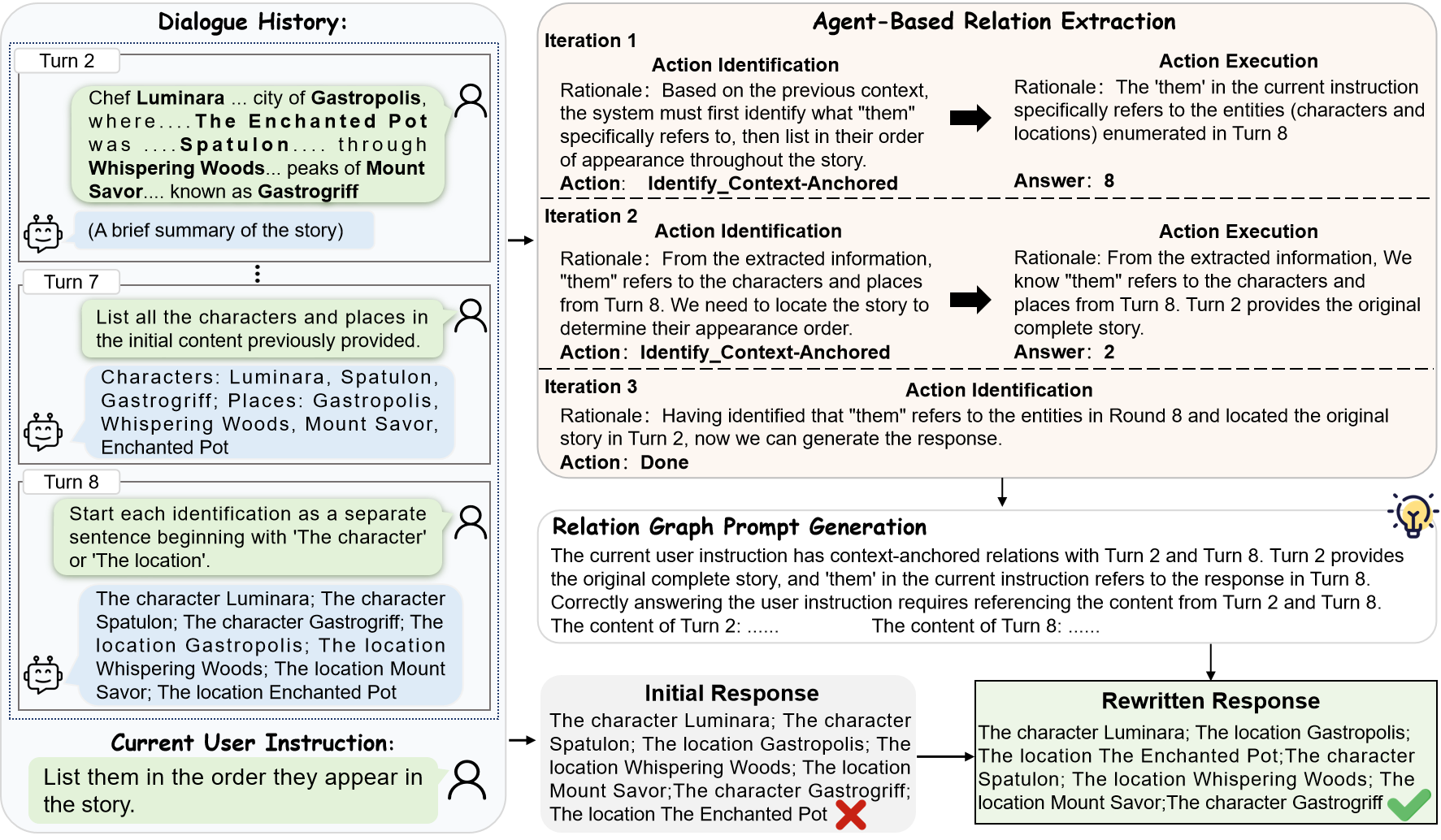}
    \caption{A typical case demonstrating how GraphIF iteratively extracts relations and corrects errors in the initial response through generated relation graph prompts.}
    \label{fig:casestudy}
\end{figure*}

\subsubsection{Limitations of Memory-Enhanced Mechanisms.}
All Memory-enhanced methods fail to significantly improve multi-turn instruction-following capabilities. MemoryBank employs coarse-grained summaries with naive RAG mechanisms that retrieve content based solely on vector similarity, missing crucial inter-turn semantic relations. MemoChat's fine-grained topic-based indexing fails to capture macro-level dialogue structure, often retrieving irrelevant information that interferes with contextual understanding.

\subsubsection{Performance Analysis Across Constraint Types.}
Figure~\ref{fig:constraint_structbench} reveals GraphIF's effectiveness across different constraint types on StructFlowBench*. While Instruction-tuned LLMs struggle particularly with global constraints and show performance drops of up to 40\%, GraphIF achieves remarkable improvements in this challenging constraint type. 
Beyond global constraints, GraphIF maintains or enhances performance across all constraint types, demonstrating its robust capability to adapt to diverse dialogue challenges. The improvement stems from GraphIF's ability to identify relevant relation types and locate specific dialogue turns that establish these relations. Results on MT-Eval* demonstrate similar experimental results. Notably, GraphIF also improves performance on current instruction constraints, demonstrating that the rewritten responses better address immediate user requirements while maintaining consistency with historical constraints.

\subsubsection{Scalability Across Model Sizes.}
Table \ref{tab:mteval} demonstrates GraphIF's effectiveness on MT-Eval* across LLMs of varying sizes. The consistent improvements across all model scales validate the generalizability and robustness of GraphIF. Similar results are observed on StructFlowBench*. The framework's plug-and-play nature allows seamless integration with any instruction-tuned LLM, offering a practical solution for enhancing multi-turn dialogue capabilities across diverse deployment scenarios without the need for model retraining or architectural changes.

\subsection{Ablation Study}
\subsubsection{The Effect of Agent-Based Relation Extraction.}
For inter-turn relation extraction, we design an iterative agent-based module that alternately performs action identification and execution. To validate its effectiveness, we replace this module with directly extracting all inter-turn relations via one-time LLM inference. As shown in Table \ref{tab:ablation}, direct LLM extraction yields significantly lower performance than our iterative approach. This degradation indicates that one-time extraction cannot adequately capture the complex inter-turn relations inherent in multi-turn dialogues, highlighting the importance of our iterative agent-based mechanism.

\subsubsection{The Effect of Relation Graph Prompt Generation.}
In the Relation Graph Prompt Generation module, we design graph prompts to explain inter-turn relations. To validate this necessity, we replace graph prompts with relation-free prompts containing only dialogue content without semantic relation explanations. Table \ref{tab:ablation} reveals significant performance degradation when using relation-free prompts. This result indicates that LLMs cannot effectively infer inter-turn semantic relations from dialogue content alone, highlighting the critical role of explicit graph prompts in guiding models to understand and satisfy inter-turn constraints.

\subsection{Case Study}
We present a specific case of Llama-3.1-8B-Instruct in Figure \ref{fig:casestudy}. The current user instruction is ``List them in the order they appear in the story", which requires reordering entities from turn 8 according to their appearance sequence in the story from turn 2. The initial response of LLM exhibits ordering confusion issues. Through iterative relation extraction, GraphIF identifies the Context-Anchored relation between the current instruction and the dialogue content of turn 2 and turn 8, and generates a graph prompt based on this relation to explain the semantic connection. Guided by the graph prompt, the LLM adjusts the entity ordering in the initial response, thereby generating the correct response. This example demonstrates that GraphIF can rectify errors in initial responses through accurate relation identification, thus improving response satisfaction for current instructions.
\section{Conclusion}
We propose GraphIF, a training-free and plug-and-play framework that models multi-turn dialogues as directed relation graphs and leverages graph prompts to enhance the instruction-following capabilities of LLMs. We carefully design two long multi-turn dialogue datasets based on public benchmarks to better evaluate the instruction-following capabilities. Extensive experiments on both datasets demonstrate that GraphIF can be seamlessly integrated into instruction-tuned LLMs, achieving significant improvements across all four evaluation metrics and maintaining consistent performance gains across different model scales. These results suggest that GraphIF provides a practical solution for enhancing multi-turn instruction following for LLMs in real-world applications.

\section{Acknowledgments}
The research was partially supported by the China National Natural Science Foundation with no. 62132018, and Hefei Key Technology Research and Development Project (2024SZD005).

\bibliography{aaai2026}

@article{achiam2023gpt,
  title={Gpt-4 technical report},
  author={Achiam, Josh and Adler, Steven and Agarwal, Sandhini and Ahmad, Lama and Akkaya, Ilge and Aleman, Florencia Leoni and Almeida, Diogo and Altenschmidt, Janko and Altman, Sam and Anadkat, Shyamal and others},
  journal={arXiv preprint arXiv:2303.08774},
  year={2023}
}

@article{dubey2024llama,
  title={The llama 3 herd of models},
  author={Dubey, Abhimanyu and Jauhri, Abhinav and Pandey, Abhinav and Kadian, Abhishek and Al-Dahle, Ahmad and Letman, Aiesha and Mathur, Akhil and Schelten, Alan and Yang, Amy and Fan, Angela and others},
  journal={arXiv e-prints},
  pages={arXiv--2407},
  year={2024}
}

@article{team2024qwen2,
  title={Qwen2 technical report},
  author={Team, Qwen},
  journal={arXiv preprint arXiv:2407.10671},
  year={2024}
}

@article{chiang2023vicuna,
  title={Vicuna: An open-source chatbot impressing gpt-4 with 90\%* chatgpt quality},
  author={Chiang, Wei-Lin and Li, Zhuohan and Lin, Ziqing and Sheng, Ying and Wu, Zhanghao and Zhang, Hao and Zheng, Lianmin and Zhuang, Siyuan and Zhuang, Yonghao and Gonzalez, Joseph E and others},
  journal={See https://vicuna. lmsys. org (accessed 14 April 2023)},
  volume={2},
  number={3},
  pages={6},
  year={2023}
}

@inproceedings{ding2023enhancing,
  title={Enhancing chat language models by scaling high-quality instructional conversations},
  author={Ding, Ning and Chen, Yulin and Xu, Bokai and Qin, Yujia and Hu, Shengding and Liu, Zhiyuan and Sun, Maosong and Zhou, Bowen},
  booktitle={Proceedings of the 2023 Conference on Empirical Methods in Natural Language Processing},
  pages={3029--3051},
  year={2023}
}

@article{zhao2024wildchat,
  title={Wildchat: 1m chatgpt interaction logs in the wild},
  author={Zhao, Wenting and Ren, Xiang and Hessel, Jack and Cardie, Claire and Choi, Yejin and Deng, Yuntian},
  journal={arXiv preprint arXiv:2405.01470},
  year={2024}
}

@inproceedings{wu2025instruct,
    title = "Review-Instruct: A Review-Driven Multi-Turn Conversations Generation Method for Large Language Models",
    author = "Wu, Jiangxu  and
      Wang, Cong  and
      Su, Tianhuang  and
      Haozhi, Lin  and
      JunYang, JunYang  and
      Zhangchao, Zhangchao  and
      Pan, Binqiang  and
      SongpanYang, SongpanYang  and
      Mingpeng, Mingpeng  and
      Shi, Kai  and
      Li, Zixian",
    booktitle = "Findings of the Association for Computational Linguistics: ACL 2025",
    year = "2025",
   
}

@article{wang2023openchat,
  title={Openchat: Advancing open-source language models with mixed-quality data},
  author={Wang, Guan and Cheng, Sijie and Zhan, Xianyuan and Li, Xiangang and Song, Sen and Liu, Yang},
  journal={arXiv preprint arXiv:2309.11235},
  year={2023}
}

@article{chen2025consistentchat,
  title={ConsistentChat: Building Skeleton-Guided Consistent Dialogues for Large Language Models from Scratch},
  author={Chen, Jiawei and Guan, Xinyan and Yuan, Qianhao and Mo, Guozhao and Zhou, Weixiang and Lu, Yaojie and Lin, Hongyu and He, Ben and Sun, Le and Han, Xianpei},
  journal={arXiv preprint arXiv:2506.03558},
  year={2025}
}

@article{gao2025strategic,
  title={A Strategic Coordination Framework of Small LLMs Matches Large LLMs in Data Synthesis},
  author={Gao, Xin and Pei, Qizhi and Tang, Zinan and Li, Yu and Lin, Honglin and Wu, Jiang and Wu, Lijun and He, Conghui},
  journal={arXiv preprint arXiv:2504.12322},
  year={2025}
}

@article{cao2025condor,
  title={Condor: Enhance llm alignment with knowledge-driven data synthesis and refinement},
  author={Cao, Maosong and Zhang, Taolin and Li, Mo and Zhang, Chuyu and Liu, Yunxin and Duan, Haodong and Zhang, Songyang and Chen, Kai},
  journal={arXiv preprint arXiv:2501.12273},
  year={2025}
}

@inproceedings{yao2023react,
  title={React: Synergizing reasoning and acting in language models},
  author={Yao, Shunyu and Zhao, Jeffrey and Yu, Dian and Du, Nan and Shafran, Izhak and Narasimhan, Karthik and Cao, Yuan},
  booktitle={International Conference on Learning Representations (ICLR)},
  year={2023}
}

@article{erdogan2025plan,
  title={Plan-and-act: Improving planning of agents for long-horizon tasks},
  author={Erdogan, Lutfi Eren and Lee, Nicholas and Kim, Sehoon and Moon, Suhong and Furuta, Hiroki and Anumanchipalli, Gopala and Keutzer, Kurt and Gholami, Amir},
  journal={arXiv preprint arXiv:2503.09572},
  year={2025}
}

@article{lambert2024tulu,
  title={Tulu 3: Pushing frontiers in open language model post-training},
  author={Lambert, Nathan and Morrison, Jacob and Pyatkin, Valentina and Huang, Shengyi and Ivison, Hamish and Brahman, Faeze and Miranda, Lester James V and Liu, Alisa and Dziri, Nouha and Lyu, Shane and others},
  journal={arXiv preprint arXiv:2411.15124},
  year={2024}
}

@article{li2025beyond,
  title={Beyond single-turn: A survey on multi-turn interactions with large language models},
  author={Li, Yubo and Shen, Xiaobin and Yao, Xinyu and Ding, Xueying and Miao, Yidi and Krishnan, Ramayya and Padman, Rema},
  journal={arXiv preprint arXiv:2504.04717},
  year={2025}
}

@article{zhang2025survey,
  title={A Survey on Multi-Turn Interaction Capabilities of Large Language Models},
  author={Zhang, Chen and Dai, Xinyi and Wu, Yaxiong and Yang, Qu and Wang, Yasheng and Tang, Ruiming and Liu, Yong},
  journal={arXiv preprint arXiv:2501.09959},
  year={2025}
}

@inproceedings{sun2024parrot,
  title={Parrot: Enhancing Multi-Turn Instruction Following for Large Language Models},
  author={Sun, Yuchong and Liu, Che and Zhou, Kun and Huang, Jinwen and Song, Ruihua and Zhao, Wayne Xin and Zhang, Fuzheng and Zhang, Di and Gai, Kun},
  booktitle={Proceedings of the 62nd Annual Meeting of the Association for Computational Linguistics (Volume 1: Long Papers)},
  pages={9729--9750},
  year={2024}
}

@article{zheng2023judging,
  title={Judging llm-as-a-judge with mt-bench and chatbot arena},
  author={Zheng, Lianmin and Chiang, Wei-Lin and Sheng, Ying and Zhuang, Siyuan and Wu, Zhanghao and Zhuang, Yonghao and Lin, Zi and Li, Zhuohan and Li, Dacheng and Xing, Eric and others},
  journal={Advances in Neural Information Processing Systems},
  volume={36},
  pages={46595--46623},
  year={2023}
}

@inproceedings{bai2024mt,
  title={MT-Bench-101: A Fine-Grained Benchmark for Evaluating Large Language Models in Multi-Turn Dialogues},
  author={Bai, Ge and Liu, Jie and Bu, Xingyuan and He, Yancheng and Liu, Jiaheng and Zhou, Zhanhui and Lin, Zhuoran and Su, Wenbo and Ge, Tiezheng and Zheng, Bo and others},
  booktitle={Proceedings of the 62nd Annual Meeting of the Association for Computational Linguistics (Volume 1: Long Papers)},
  pages={7421--7454},
  year={2024}
}

@inproceedings{kwan2024mt,
  title={MT-Eval: A Multi-Turn Capabilities Evaluation Benchmark for Large Language Models},
  author={Kwan, Wai-Chung and Zeng, Xingshan and Jiang, Yuxin and Wang, Yufei and Li, Liangyou and Shang, Lifeng and Jiang, Xin and Liu, Qun and Wong, Kam-Fai},
  booktitle={Proceedings of the 2024 Conference on Empirical Methods in Natural Language Processing},
  pages={20153--20177},
  year={2024}
}

@article{he2024multi,
  title={Multi-if: Benchmarking llms on multi-turn and multilingual instructions following},
  author={He, Yun and Jin, Di and Wang, Chaoqi and Bi, Chloe and Mandyam, Karishma and Zhang, Hejia and Zhu, Chen and Li, Ning and Xu, Tengyu and Lv, Hongjiang and others},
  journal={arXiv preprint arXiv:2410.15553},
  year={2024}
}

@article{li2025structflowbench,
  title={Structflowbench: A structured flow benchmark for multi-turn instruction following},
  author={Li, Jinnan and Li, Jinzhe and Wang, Yue and Chang, Yi and Wu, Yuan},
  journal={arXiv preprint arXiv:2502.14494},
  year={2025}
}

@article{edge2024local,
  title={From local to global: A graph rag approach to query-focused summarization},
  author={Edge, Darren and Trinh, Ha and Cheng, Newman and Bradley, Joshua and Chao, Alex and Mody, Apurva and Truitt, Steven and Metropolitansky, Dasha and Ness, Robert Osazuwa and Larson, Jonathan},
  journal={arXiv preprint arXiv:2404.16130},
  year={2024}
}

@article{guo2024lightrag,
  title={LightRAG: Simple and Fast Retrieval-Augmented Generation},
  author={Guo, Zirui and Xia, Lianghao and Yu, Yanhua and Ao, Tu and Huang, Chao},
  journal={arXiv preprint arXiv:2410.05779},
  year={2024}
}

@inproceedings{wang2024knowledge,
  title={Knowledge graph prompting for multi-document question answering},
  author={Wang, Yu and Lipka, Nedim and Rossi, Ryan A and Siu, Alexa and Zhang, Ruiyi and Derr, Tyler},
  booktitle={Proceedings of the AAAI Conference on Artificial Intelligence},
  volume={38},
  pages={19206--19214},
  year={2024}
}

@inproceedings{li2024graphreader,
  title={GraphReader: Building Graph-based Agent to Enhance Long-Context Abilities of Large Language Models},
  author={Li, Shilong and He, Yancheng and Guo, Hangyu and Bu, Xingyuan and Bai, Ge and Liu, Jie and Liu, Jiaheng and Qu, Xingwei and Li, Yangguang and Ouyang, Wanli and others},
  booktitle={Findings of the Association for Computational Linguistics: EMNLP 2024},
  pages={12758--12786},
  year={2024}
}

@inproceedings{lin2025rje,
  title={RJE: A Retrieval-Judgment-Exploration Framework for Efficient Knowledge Graph Question Answering with LLMs},
  author={Lin, Can and Jiang, Zhengwang and Zheng, Ling and Zhao, Qi and Zhang, Yuhang and Song, Qi and Zhou, Wangqiu},
  booktitle={Proceedings of the 2025 Conference on Empirical Methods in Natural Language Processing},
  pages={17288--17305},
  year={2025}
}

@article{jimenez2024hipporag,
  title={Hipporag: Neurobiologically inspired long-term memory for large language models},
  author={Jimenez Gutierrez, Bernal and Shu, Yiheng and Gu, Yu and Yasunaga, Michihiro and Su, Yu},
  journal={Advances in Neural Information Processing Systems},
  volume={37},
  pages={59532--59569},
  year={2024}
}

@inproceedings{zhong2024memorybank,
  title={Memorybank: Enhancing large language models with long-term memory},
  author={Zhong, Wanjun and Guo, Lianghong and Gao, Qiqi and Ye, He and Wang, Yanlin},
  booktitle={Proceedings of the AAAI Conference on Artificial Intelligence},

  pages={19724--19731},
  year={2024}
}

@article{lu2023memochat,
  title={Memochat: Tuning llms to use memos for consistent long-range open-domain conversation},
  author={Lu, Junru and An, Siyu and Lin, Mingbao and Pergola, Gabriele and He, Yulan and Yin, Di and Sun, Xing and Wu, Yunsheng},
  journal={arXiv preprint arXiv:2308.08239},
  year={2023}
}

@article{zhang2024cfbench,
  title={Cfbench: A comprehensive constraints-following benchmark for llms},
  author={Zhang, Tao and Zhu, Chenglin and Shen, Yanjun and Luo, Wenjing and Zhang, Yan and Liang, Hao and Yang, Fan and Lin, Mingan and Qiao, Yujing and Chen, Weipeng and others},
  journal={arXiv preprint arXiv:2408.01122},
  year={2024}
}

@article{qin2024infobench,
  title={Infobench: Evaluating instruction following ability in large language models},
  author={Qin, Yiwei and Song, Kaiqiang and Hu, Yebowen and Yao, Wenlin and Cho, Sangwoo and Wang, Xiaoyang and Wu, Xuansheng and Liu, Fei and Liu, Pengfei and Yu, Dong},
  journal={arXiv preprint arXiv:2401.03601},
  year={2024}
}

\clearpage
\onecolumn
\section{Appendix}

\subsection{Overview}
We first provide a detailed introduction to the relations defined in GraphIF and their implementation, then present a comprehensive description of the construction methods and statistics of our two constructed datasets, followed by an introduction to the baseline implementation. Subsequently, we present the experimental evaluation section, including evaluation metrics and experimental validation of LLM-Judge effectiveness. We then conduct Performance Analysis Across Constraint Types on MT-Eval* and Scalability Across Model Sizes of GraphIF on StructFlowBench*. Furthermore, we provide an analysis of the latency efficiency of GraphIF, followed by its performance on the recently released public benchmark MultiChallenge, the performance of superior LLMs, and then a comparison with two additional baselines. Finally, we provide the prompt templates used by GraphIF.
\subsection{Relations in GraphIF and Implementation}
We define the following actions, where each action represents an inter-turn relation, and introduce the specific implementation functions for each action:
\begin{itemize}
    \item \textit{Identify\_Global\_ Constraint}: Identifies the current instruction that serves as a global constraint for subsequent interactions. For example, the current instruction is ``All subsequent responses should end with Is there anything else I can help you with?'' or ``From now on, don't use any commas in your answers". In the implementation, we record a Global Constraint Set, which contains all global constraints. We first determine whether the current instruction conflicts with or updates any constraint in the Global Constraint Set, for example, when a previous dialogue turn requires all subsequent responses to begin with letter `A' while the current instruction requires beginning with letter `T'. If conflicts or updates exist, the latest constraint takes precedence and updates are made accordingly; otherwise, the current instruction is directly added to the Global Constraint Set, and all subsequent instructions will be connected to the relevant dialogue turns in the Global Constraint Set via directed edges labeled \textit{Global Constraint}.
    \item \textit{Identify\_Context\_Anchored}: Identifies specific historical dialogue content that the instruction semantically relies on or logically connects to. For example, the current instruction is ``Based on the initial content,...", which logically requires identifying the content referred to by the initial content. In the implementation, we iteratively locate the most relevant unrecorded dialogue turn that exhibits a Context Anchored relation based on already extracted information. For instance, when the current user instruction semantically establishes Context Anchored relations with multiple historical conversational turns, we employ an iterative extraction approach, identifying the most relevant conversational turn per iteration to achieve more accurate relation extraction.
    \item \textit{Identify\_Modify}: Identifies specific historical dialogue content that the current instruction refines, changes, modifies, or extends. Modify relation is built upon context-anchored relations with additional requirements. For example, the current instruction is ``Rewrite it in the style of a formal scientific report", which further imposes new requirements that necessitate modifying the previous response. The implementation follows the same approach as Context Anchored relation extraction, iteratively locating the current unrecorded dialogue turn that best satisfies the Modify relation.
     \item \textit{Identify\_Summary}: Identifies Summary relation, which indicates that the current instruction explicitly or implicitly requests a summary of specific historical round(s) in the dialogue history. In the implementation, we prompt the large language model to return all historical dialogue turn IDs that the current instruction requires summarizing based on the conversation history and current user instruction.
    \item \textit{New\_Topic}: Identifies  New\_Topic relation, which indicates that the current instruction  introduces entirely new topics unconnected to the current dialogue context. In the implementation, we record the dialogue turns contained within each topic and maintain a topic pointer that indicates the current topic. During GraphIF execution, only the historical dialogue content within the current topic is analyzed while filtering out irrelevant historical conversational content. When a topic shift occurs, we first determine whether the current user instruction belongs to any existing historical topic. If so, the topic pointer jumps to the corresponding topic; otherwise, a new topic is initiated if the instruction is determined not to belong to any historical topic.
    \item \textit{Done}: Indicates complete relation extraction termination.
\end{itemize}
Unlike the relations defined in previous benchmarks, we approach inter-turn relations from the perspective of semantic relational structures, providing a more inclusive and coarse-grained categorization that encompasses all previously defined fine-grained relations. This semantic relation-based definition is more aligned with our relation extraction task, improving the accuracy of relation identification.

The set of relations in GraphIF, while refined from existing benchmarks, is designed to be extensible rather than exhaustive. This allows the framework to seamlessly incorporate new relations in the future. As a design principle, each relation should be equipped with a precise semantic explanation to enhance the LLM's understanding.

\subsection{Dataset}
We introduce the process of constructing the datasets MT-Eval* and StructFlowBench*. Table~\ref{tab:statistics} shows the statistics of two datasets.
\begin{table*}[ht]
\centering
\begin{tabular}{ccccc}
\toprule
Dataset & Sample Size & Turns per sample & Total Dialogue Nums & Relation Nums  \\
\midrule
MT-Eval* & 10 & 23 & 10*23=230 & 4 \\
StructFlowBench* & 32 & 24 & 32*24=768 & 7\\
\bottomrule
\end{tabular}
\caption{Statistics of MT-Eval* and StructFlowBench*.}
\label{tab:statistics}
\end{table*}
\begin{itemize}
    \item \textbf{MT-Eval*} We construct MT-Eval* based on MT-Eval. MT-Eval designs prompts based on document content to enable LLMs to conduct question-answering based on the document, thereby generating multi-turn conversations. However, MT-Eval generates multiple multi-turn conversation samples for a single document, where each sample has very few turns and contains only one type of inter-turn relation, leading to oversimplified inter-turn relation modeling that cannot accurately represent real-world scenarios. Therefore, we integrate and reorganize the multiple samples generated from the same document in MT-Eval and conduct manual inspection and modification, constructing a long multi-turn conversation dataset, MT-Eval*, which contains 10 samples with 23 dialogue turns each.

    Figure~\ref{fig:relation-mteval} shows the dialogue relation graph of samples in MT-Eval*. The first turn establishes a global instruction that all subsequent responses must satisfy, therefore creating directed edges of the Global Constraint type between it and Turns 2-23. Turn 2 and Turn 19 respectively, introduce two unrelated stories, thus Turn 19 has a \textit{New\_topic} relation with the previous context, indicating that Turn 19 initiates an entirely new topic. For visual clarity, we omit these edges in the figure.
\begin{figure}[ht]
    \centering
    \includegraphics[width=0.75\linewidth]{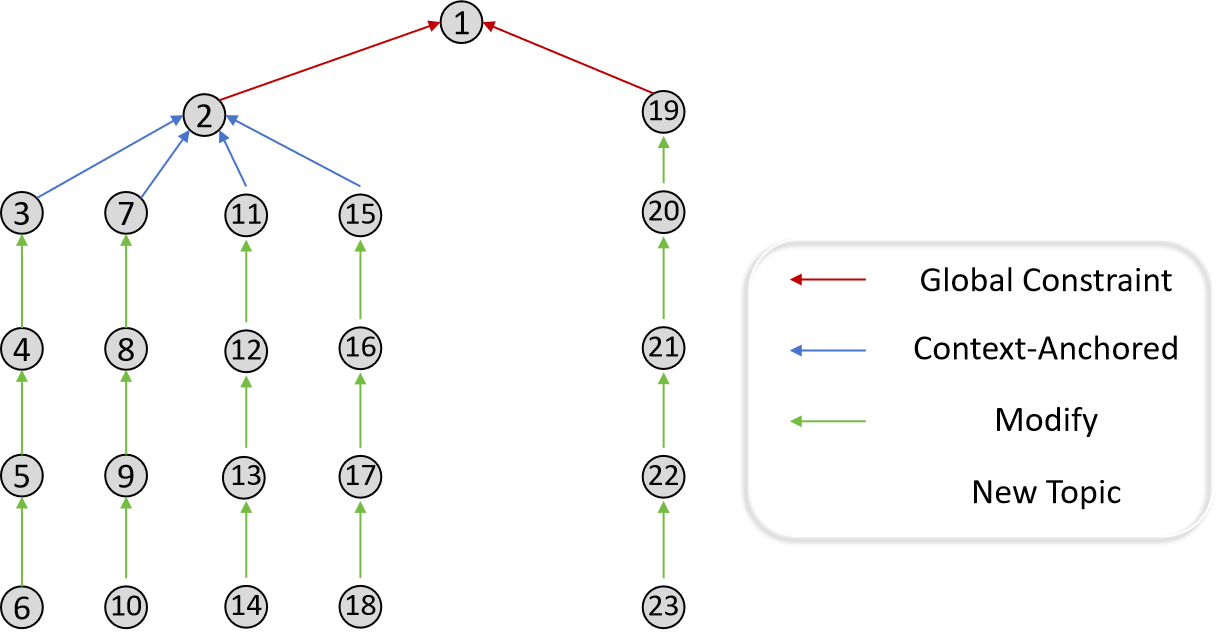}
    \caption{Relation graph of MT-Eval*}
    \label{fig:relation-mteval}
\end{figure}

    \item \textbf{StructFlowBench*} We construct StructFlowBench* based on StructFlowBench. StructFlowBench supports customizable inter-turn relations and generates multi-turn dialogue datasets using LLMs according to the configured relations. Given that the official StructFlowBench dataset suffers from the same limitations as MT-Eval—short dialogue turns with each sample containing only one type of inter-turn relation—we designed a complex relation structure involving 24 dialogue turns. Using StructFlowBench's prompts and GPT-4o, we generated multi-turn dialogue data, followed by manual inspection and correction, resulting in StructFlowBench*: a long multi-turn conversation
    dataset containing 32 samples of 24 turns each.
    Figure~\ref{fig:relation-structbench} shows the dialogue relation graph of samples in StructFlowBench*.  Turn 5 and Turn 17 both impose global constraints on all subsequent conversations, therefore Global Constraint relations exist between Turn 5 and Turns 6-24 (similarly for Turn 17 and Turns 18-24). Turn 13 requires summarizing Turns 1-12, while Turn 24 requires summarizing Turns 14-23. Turn 1 and Turn 14 each initiate a topic, therefore a New Topic relation exists between Turn 14 and Turns 1-13. For visual clarity, we omit these edges in the figure. For detailed definitions of each constraint type, please refer to StructFlowBench.

    \begin{figure}[ht]
        \centering
        \includegraphics[width=0.75\linewidth]{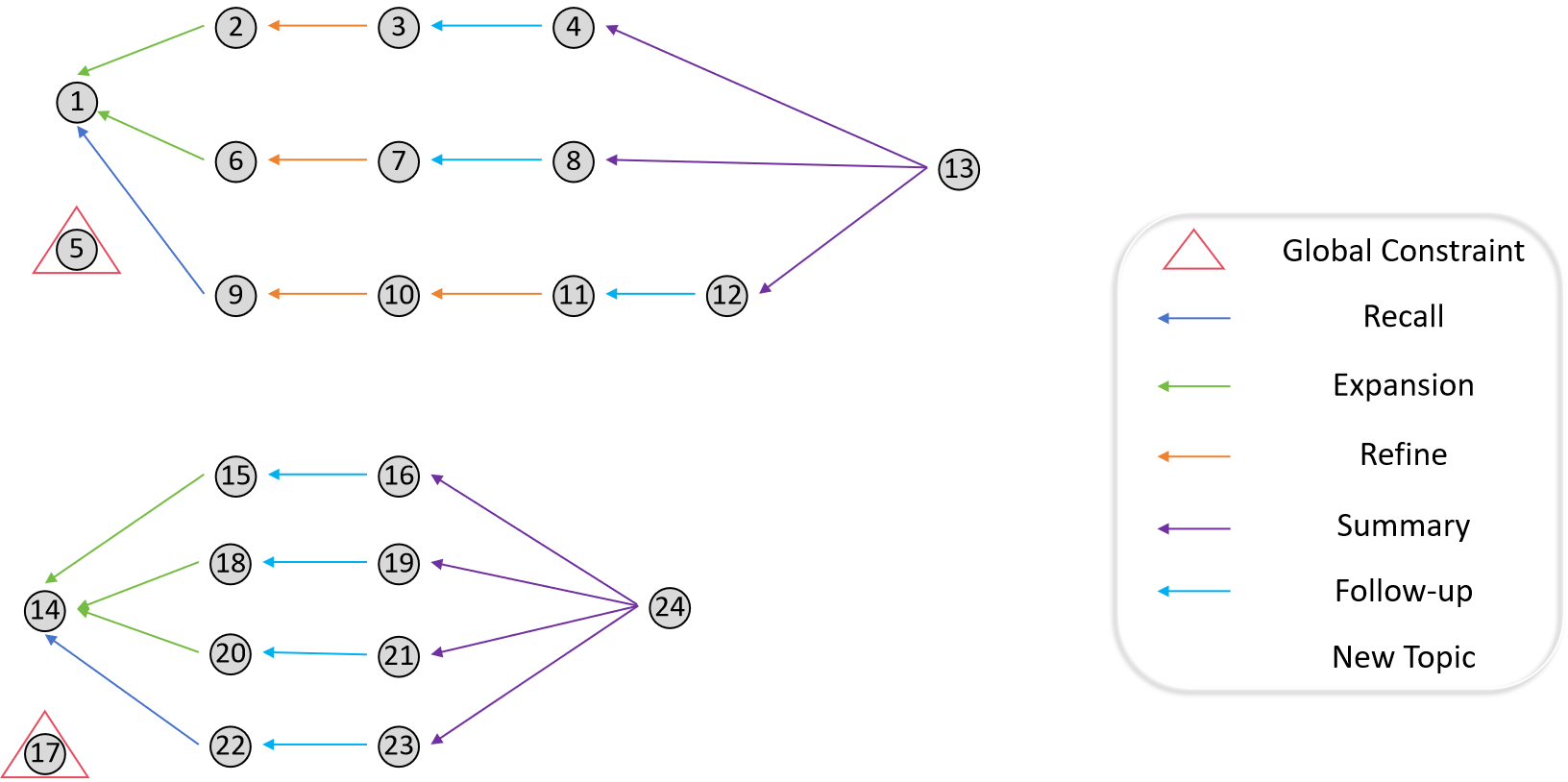}
        \caption{Relation graph of StructFlowBench*}
        \label{fig:relation-structbench}
    \end{figure}
\end{itemize}

\subsection{Baseline}
\begin{itemize}
    \item \textbf{MemoryBank}: To better adapt to the multi-turn instruction following task, we made minor adjustments to the official MemoryBank code: the official implementation generates both an overall summary of the historical conversation and an overall personality to summarize the user's character traits. However, our task does not involve the analysis of user personality. Therefore, we replaced the overall personality with overall constraints, which extract all constraints from the historical dialogues that need to be satisfied when responding to the current user instruction, thereby better adapting to the multi-turn instruction following task. For fair comparison, we did not fine-tune the LLMs.

    \item \textbf{MemoChat}: We directly used the official code for experiments. For fair comparison, we did not fine-tune the LLMs.
\end{itemize}

\subsection{Evaluation}
\subsubsection{Evaluation Metrics}
We employ four evaluation metrics for multi-turn instruction following, following the definitions in StructFlowBench.
\begin{itemize}
    \item \textbf{Constraint Satisfaction Rate (CSR)} evaluates the average proportion of satisfied constraints across all instructions:
    \begin{equation}
        CSR=\frac{1}{m} \sum_{i=1}^{m} \left( \frac{1}{n_i} \sum_{j=1}^{n_i} s_j^i \right),
    \end{equation}
    where $m$ represents the total number of instructions, $n_i$ denotes the the number of constraints in the $i$-th instruction, and $s_i^j \in \{0,1\}$ indicates whether the $j$-th constraint in the $i$-th instruction is satisfied. 

    \item \textbf{Instruction Satisfaction Rate (ISR)} measures the proportion of instructions where all constraints are fully satisfied, computed as
    \begin{equation}
        ISR=\frac{1}{m} \sum_{i=1}^{m} s_i,
    \end{equation}
    where $s_i \in \{0,1\}$ indicates whether all constraints in the $i$-th instruction are satisfied.

    \item \textbf{Decomposed Requirements Following Ratio (DRFR)} evaluates the overall satisfaction of requirements across all instructions, defined as
    \begin{equation}
        DRFR=\frac{\sum_{i,j} r_{i,j}'}{\sum_i m_i},
    \end{equation}
    where $m_i$ is the number of scoring questions for the $i$-th instruction, and $r_{i,j}'$ denotes the result of the $j$-th scoring question in the $i$-th instruction.

    \item \textbf{Weighted Constraint Satisfaction Rate (WCSR)} incorporates weighted factors to account for the varying significance of different constraint types:
    \begin{equation}
        WCSR=\frac{\sum_{j=1}^{n} w_j s_j}{\sum_{j=1}^{n} w_j},
    \end{equation}
    where n denotes the total number of constraints, $w_j$ represents the weight assigned to the j-th constraint, and $s_j \in \{0,1\}$ indicates whether the $j$-th constraint is satisfied. We assign the weight of intra-turn constraints (i.e., current instruction type) as 1 ($w_r=1$), and inter-turn constraints are given a higher weight of $w_s=2$
\end{itemize}
\subsubsection{Human Verification of LLM-Judge}
We conducted experiments to validate that GPT-4o ratings can align with human ratings. Specifically, we selected 6 data samples from Llama-3.1-8B-Instruct on the MT-Eval* dataset, which contains 138 dialogue turns, and selected 10 data samples from Llama-3.1-8B-Instruct on the StructFlowBench*, which contains 240 conversational turns. We performed careful manual evaluation on the selected data and calculated the Kappa coefficient between the human evaluation results and GPT-4o evaluation results. As shown in Table~\ref{tab:kappa}, the Kappa coefficients for both datasets exceed 0.6, indicating high consistency between GPT-4o ratings and human ratings. The Kappa coefficient on the MT-Eval* dataset is notably higher than that on StructFlowBench*, which can be attributed to the simpler multi-turn instruction following constraints and shorter dialogue context length in MT-Eval*, resulting in higher accuracy of large language model evaluation. The experimental results demonstrate that utilizing advanced large language models such as GPT-4o to evaluate the output quality of other models is a reliable approach that can effectively reduce subjective bias and decrease the time costs associated with relying solely on manual evaluation. By combining LLM-Judge with human validation, we can rapidly obtain effective and accurate evaluation results.

\begin{table*}[ht]
\centering
\begin{tabular}{cc}
\toprule
Dataset & Kappa coefficient  \\
\midrule
MT-Eval* & 0.92 \\
StructFlowBench* & 0.72\\
\bottomrule
\end{tabular}
\caption{Kappa coefficient on MT-Eval* and StructFlowBench*}
\label{tab:kappa}
\end{table*}

\subsection{Performance Analysis Across Constraint Types on MT-Eval*}
\begin{figure}[h]
    \centering
    \includegraphics[width=0.7\linewidth]{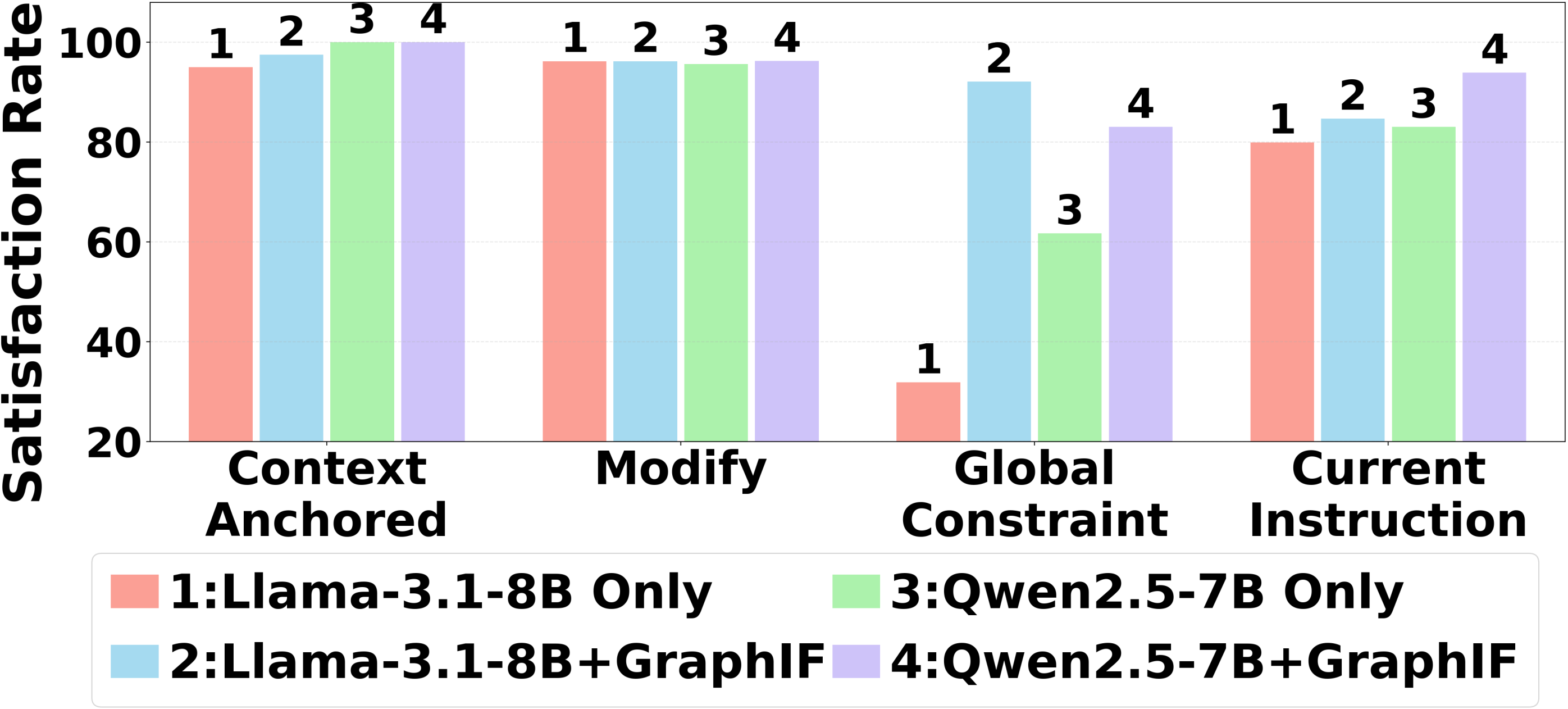}
    \caption{Detailed constraint satisfaction results across different constraint types on the MT-Eval*.}
    \label{fig:constraint_mteval}
\end{figure}

Figure~\ref{fig:constraint_mteval} demonstrates  GraphIF’s effectiveness across different constraint types on MT-Eval*, revealing consistent results with those observed on StructFlowBench*: GraphIF substantially enhances large language model performance on Global Constraint type constraints. GraphIF maintains or enhances performance across all constraint types.

\subsection{Scalability Across Model Sizes of GraphIF on StructFlowBench*}
Table~\ref{tab:structflowbench} presents the performance comparison of Qwen2.5 models of different scales as backbone models on the StructFlowBench*, which demonstrates the same results as the MT-Eval* dataset: The consistent improvements across model scales validate the generalizability and robustness of GraphIF.
\begin{table}[htbp]
    \centering
    \begin{tabular}{cccccc}
        \toprule
        \textbf{Model} & \textbf{Method} & \multicolumn{4}{c}{\textbf{StructFlowBench*(\%)}} \\
        \cmidrule(lr){3-6}
        & &  \textbf{CSR} & \textbf{ISR} & \textbf{DRFR} & \textbf{WCSR}  \\
        \midrule
        \multirow{2}{*}{\makecell[c]{\textbf{Qwen2.5-3B}\\\textbf{-Instruct}}}
        & LLM-Only & 67.42 & 20.23 & 67.88 & 63.61\\
        & +GraphIF & \textbf{75.54} & \textbf{42.95} & \textbf{76.59} & \textbf{73.07}\\
        \midrule
        \multirow{2}{*}{\makecell[c]{\textbf{Qwen2.5-7B}\\\textbf{-Instruct}}}
        & LLM-Only & 70.70 & 23.02 & 71.22 & 67.07\\
        & +GraphIF & \textbf{89.46} & \textbf{69.25} & \textbf{89.47} & \textbf{88.60}\\
        \midrule
        \multirow{2}{*}{\makecell[c]{\textbf{Qwen2.5-14B}\\\textbf{-Instruct}}}
        & LLM-Only & 74.49 & 31.64 & 75.36 & 71.09\\
        & +GraphIF & \textbf{91.74} & \textbf{75.39} & \textbf{91.53} & \textbf{90.61}\\
        \bottomrule
    \end{tabular}
    \caption{Performance comparison of Qwen2.5 models of different scales as backbone models on the StructFlowBench*}
    \label{tab:structflowbench}
\end{table}

\subsection{Latency Efficiency Analysis of GraphIF}
Table \ref{tab:latency} shows the latency efficiency of GraphIF on Qwen2.5-7B-Instruct. It is demonstrated that the substantial performance enhancement facilitated by GraphIF is achieved with a commensurate and acceptable overhead, indicating a favorable performance-overhead trade-off.

\begin{table}[ht]
    \centering
    \begin{tabular}{cccc|cc}
        \toprule
        \textbf{Model} & \textbf{Method} & \multicolumn{2}{c|}{\textbf{MT-Eval*(\%)}} & \multicolumn{2}{c}{\textbf{StructFlowBench*(\%)}}\\
        & & \textbf{Avg Calls} & \textbf{Avg Latency(s)} & \textbf{Avg Calls} & \textbf{Avg Latency(s)}\\
        
        \midrule
        \multirow{2}{*}{\makecell[c]{\textbf{Qwen2.5-7B}\\\textbf{-Instruct}}}
        & LLM-Only & 1 & 3.01 & 1 & 7.78\\
        & +GraphIF & 4.94 & 10.89 & 4.96 & 19.56 \\

        \midrule
        \multirow{2}{*}{\makecell[c]{\textbf{Llama-3.1-8B}\\\textbf{-Instruct}}}
        & LLM-Only & 1 & 3.01 & 1 & 7.78\\
        & +GraphIF & 4.94 & 10.89 & 4.96 & 19.56 \\
        
        \bottomrule
    \end{tabular}
    \caption{Latency efficiency of Qwen2.5-7B-Instruct-only vs. Qwen2.5-7B-Instruct+GraphIF on MT-Eval* and StructFlowBench*.}
    \label{tab:latency}
\end{table}

\subsection{Performance on MultiChallenge}
Table \ref{tab:multichallenge} shows the performance of GraphIF on the recently released public benchmark MultiChallenge. Overall, Qwen2.5-14B with GraphIF achieves a score of 24.02\%, compared to 17.12\% for the model without it. This result not only shows a significant improvement but also surpasses the performance of Llama-3.3-70B-Instruct reported in the MultiChallenge paper, thus validating the strong generalization capability of GraphIF.
\begin{table}[h!]
    \centering
    \small
    \begin{tabular}{ccccccc}
        \toprule
        \textbf{Model} & \textbf{Method} & \multicolumn{5}{c}{\textbf{MultiChallenge(\%)}} \\
        \cmidrule(lr){3-7}
        & &  \textbf{IM} & \textbf{SC} & \textbf{IR} & \textbf{RVE} & \textbf{Overall Score}  \\
        \midrule
        \multirow{2}{*}{\makecell[c]{\textbf{Qwen2.5-14B}\\\textbf{-Instruct}}}
        & LLM-Only & 17.70 & 10.00 & 18.84 & \textbf{21.95} & 17.12\\
        & +GraphIF & \textbf{20.35} & \textbf{20.00} & \textbf{36.23} & 19.51 & \textbf{24.02}\\
        
        \bottomrule
    \end{tabular}
    \caption{Performance of GraphIF on the recently released public benchmark MultiChallenge. IM=Inference Memory, SC=Self Coherence, IR=Instruction Retention, RVE=Reliable Version Editting.}
    \label{tab:multichallenge}
\end{table}

\subsection{Performance of Superior LLMs}
Table \ref{tab:superiorllm} shows the performance of two superior LLMs on MT-Eval* and StructFlowBench*. The experimental results demonstrate that GraphIF consistently improves performance across model scales, enhancing the capabilities of both Llama-3.1-70B (a large-scale open-source LLM) and GPT-4o-mini (a commercial LLM) across all four metrics. Notably, Llama3.1-8B+GraphIF outperforms Llama3.1-70B on both MT-Eval* and StructFlowBench*.

\begin{table*}[htbp]
    \centering
    \small
    \begin{tabular}{c|c|cccc|cccc}
        \toprule
        \multirow{2}{*}{\textbf{Backbone Model}}  &  \multirow{2}{*}{\textbf{Method}} & \multicolumn{4}{c|}{\textbf{MT-Eval*(\%)}} & \multicolumn{4}{c}{\textbf{StructFlowBench*(\%)}}\\
        & &  \textbf{CSR} & \textbf{ISR} & \textbf{DRFR} & \textbf{WCSR} & \textbf{CSR} & \textbf{ISR} & \textbf{DRFR} & \textbf{WCSR} \\
        
        \midrule
        \multirow{2}{*}{Llama-3.1-70B-Instruct} & LLM-only  & 83.99 & 60.70 & 83.56 & 83.99 & 87.18 & 64.72 & 87.74 & 87.18 \\
        & +GraphIF & \textbf{95.56} & \textbf{87.77} & \textbf{95.43} & \textbf{96.89} & \textbf{94.25} & \textbf{81.39} & \textbf{94.10} & \textbf{94.25} \\

        \midrule
        \multirow{2}{*}{GPT-4o-mini} & LLM-only  & 90.65 & 74.78 & 90.30 & 93.86 & 79.19 & 41.93 & 81.17 & 79.19 \\
        & +GraphIF & \textbf{93.03} & \textbf{79.89} & \textbf{92.80} & \textbf{95.58} & \textbf{91.16} & \textbf{77.60} & \textbf{93.38} & \textbf{91.16} \\

        \bottomrule
    \end{tabular}
    \caption{Performance of Llama-3.1-70B-Instruct and GPT-4o-mini on MT-Eval* and StructFlowBench*.}
    \label{tab:superiorllm}
\end{table*}

\subsection{The Comparison with Other Baselines}
Table \ref{tab:otherbaseline} shows the comparison with another two baselines: 
1) \textbf{MemAgent-7B}, which employs an RL-based training approach to enhance the long-context processing capabilities of LLMs; and 2) \textbf{Self-Refine}, which improves initial outputs through iterative feedback and refinement. 

Our experimental results, however, found that neither MemAgent nor Self-Refine yielded improvement on our datasets when using Qwen2.5-7B. The potential reasons are: MemAgent targets long-context scenarios primarily through memory enhancement, without a specific focus on augmenting relational understanding; whereas Self-Refine, while relying on self-correction mechanisms, is susceptible to comprehension errors during its iterative process.

\begin{table*}[htbp]
    \centering
    \small
    \begin{tabular}{c|cccc|cccc}
        \toprule
        \textbf{Method} & \multicolumn{4}{c|}{\textbf{MT-Eval*(\%)}} & \multicolumn{4}{c}{\textbf{StructFlowBench*(\%)}}\\
        &   \textbf{CSR} & \textbf{ISR} & \textbf{DRFR} & \textbf{WCSR} & \textbf{CSR} & \textbf{ISR} & \textbf{DRFR} & \textbf{WCSR} \\
        
        \midrule
        Qwen2.5-7B-only  & 80.22 & 51.30 & 79.70 & 79.57 & 70.70 & 23.02 & 71.22 & 67.07 \\
        \midrule
        MemAgent-7B  & 65.72 & 25.22 & 65.15 & 65.10 & 75.56 & 32.81 & 76.06 & 75.56 \\
        \midrule
         Qwen2.5-7B+Self-Refine& 68.70 & 26.52 & 68.03 & 65.62 & 74.75 & 29.95 & 76.28 & 74.75 \\
        \midrule
         Qwen2.5-7B+GraphIF & \textbf{91.30} & \textbf{76.96} & \textbf{91.06} & \textbf{90.72} & \textbf{89.46} & \textbf{69.25} & \textbf{89.47} & \textbf{88.60} \\

        \bottomrule
    \end{tabular}
    \caption{Performance comparison with another two baselines: MemAgent-7B and Qwen2.5-7B+Self-Refine.}
    \label{tab:otherbaseline}
\end{table*}

\subsection{Prompt Template}
\begin{figure}[htbp]
    \centering
    \includegraphics[width=1\linewidth]{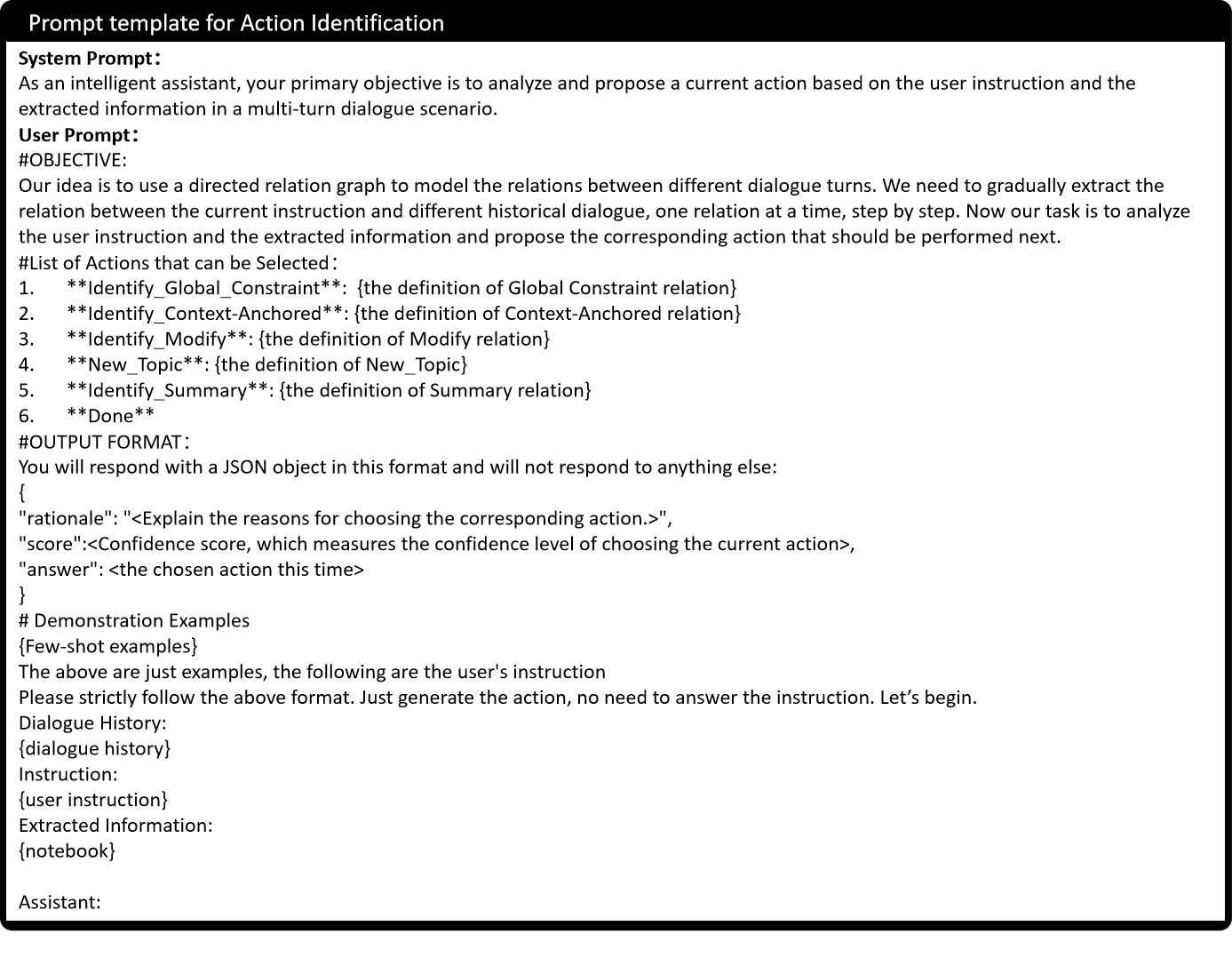}
    \caption{Prompt template for Action Identification}
    \label{fig:action_identification}
\end{figure}

\begin{figure}[htbp]
    \centering
    \includegraphics[width=1.0\linewidth]{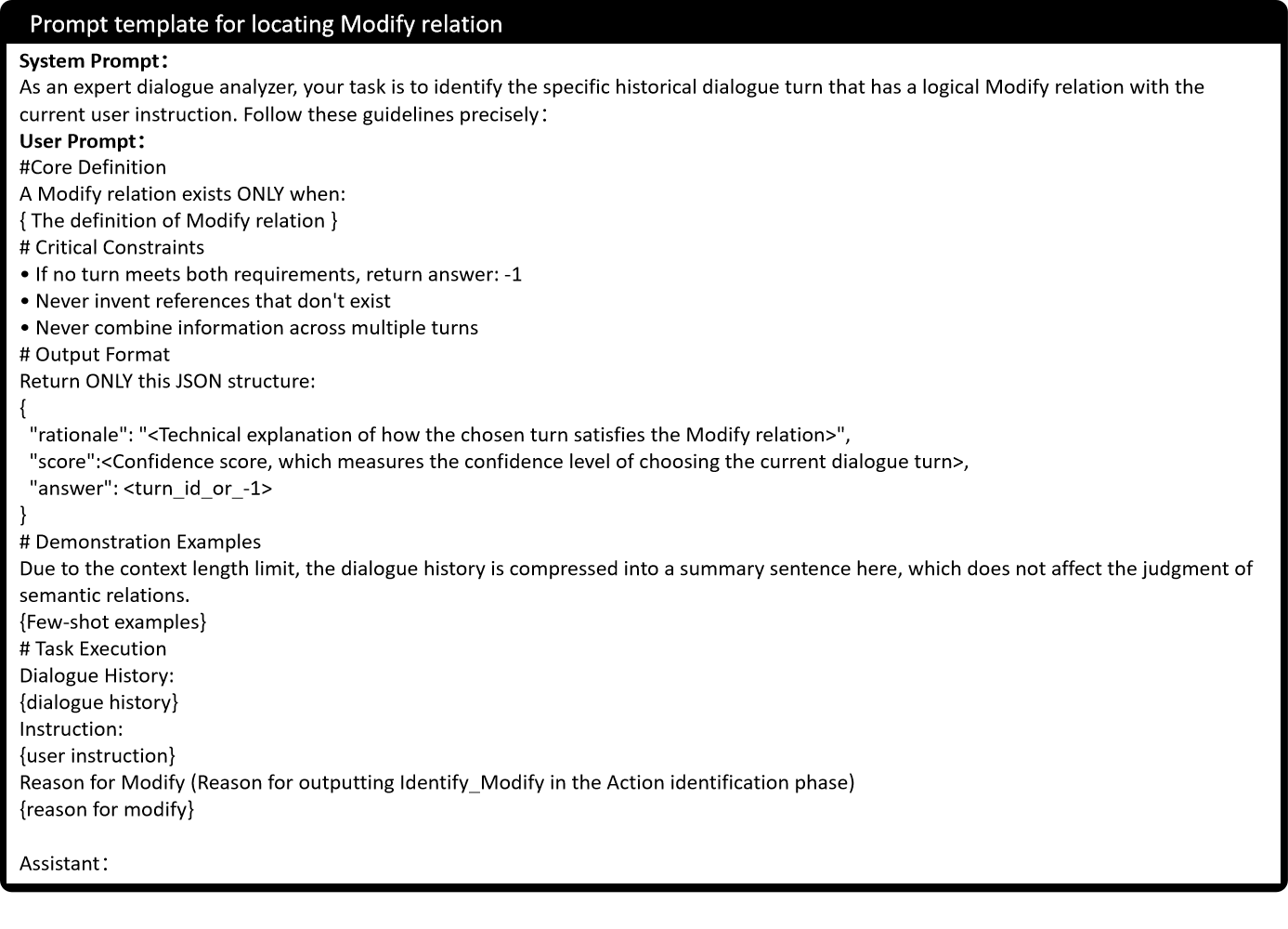}
    \caption{Prompt template for locating Modify relation. The same applies to Context-Anchored relation.}
    \label{fig:identify_modify}
\end{figure}

\begin{figure}[htbp]
    \centering
    \includegraphics[width=1\linewidth]{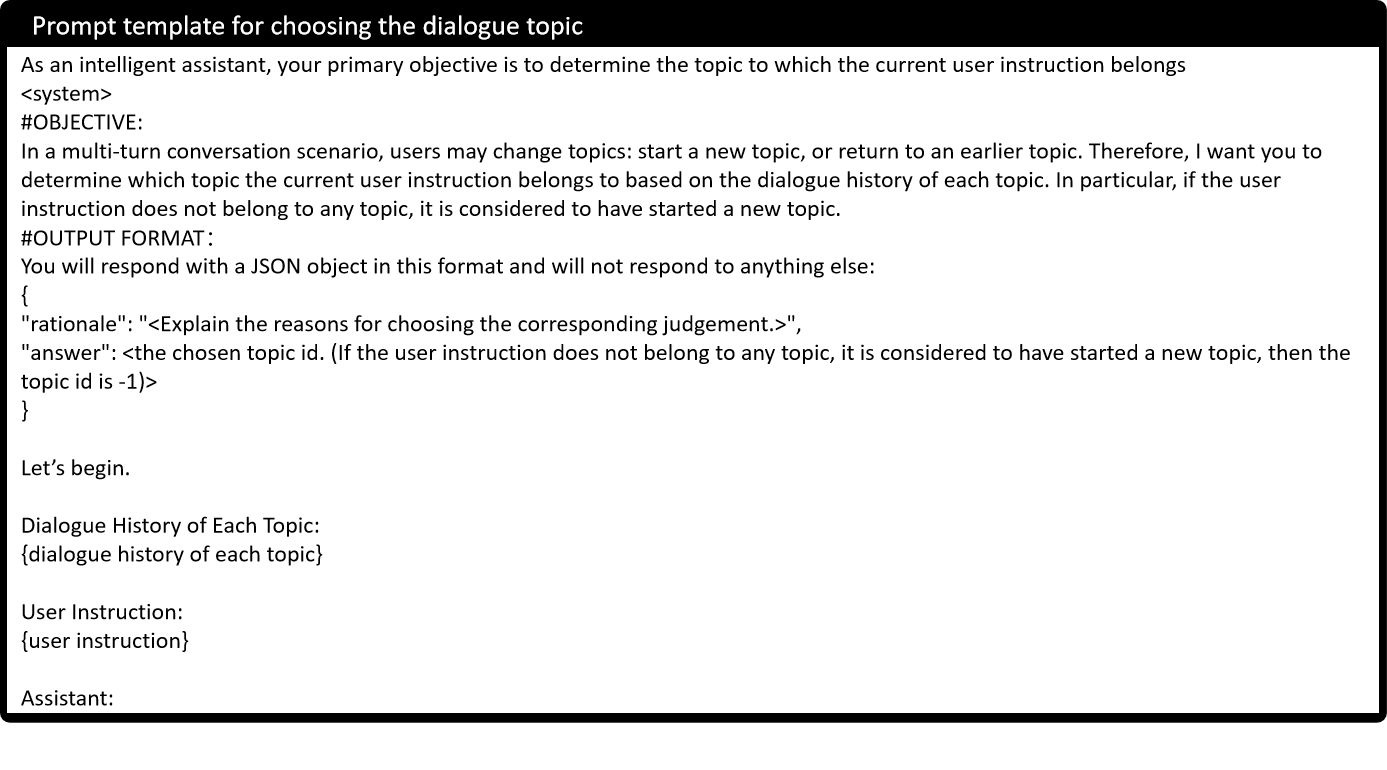}
    \caption{Prompt template for choosing the dialogue topic.}
    \label{fig:choose_topic}
\end{figure}

\begin{figure}[htbp]
    \centering
    \includegraphics[width=1\linewidth]{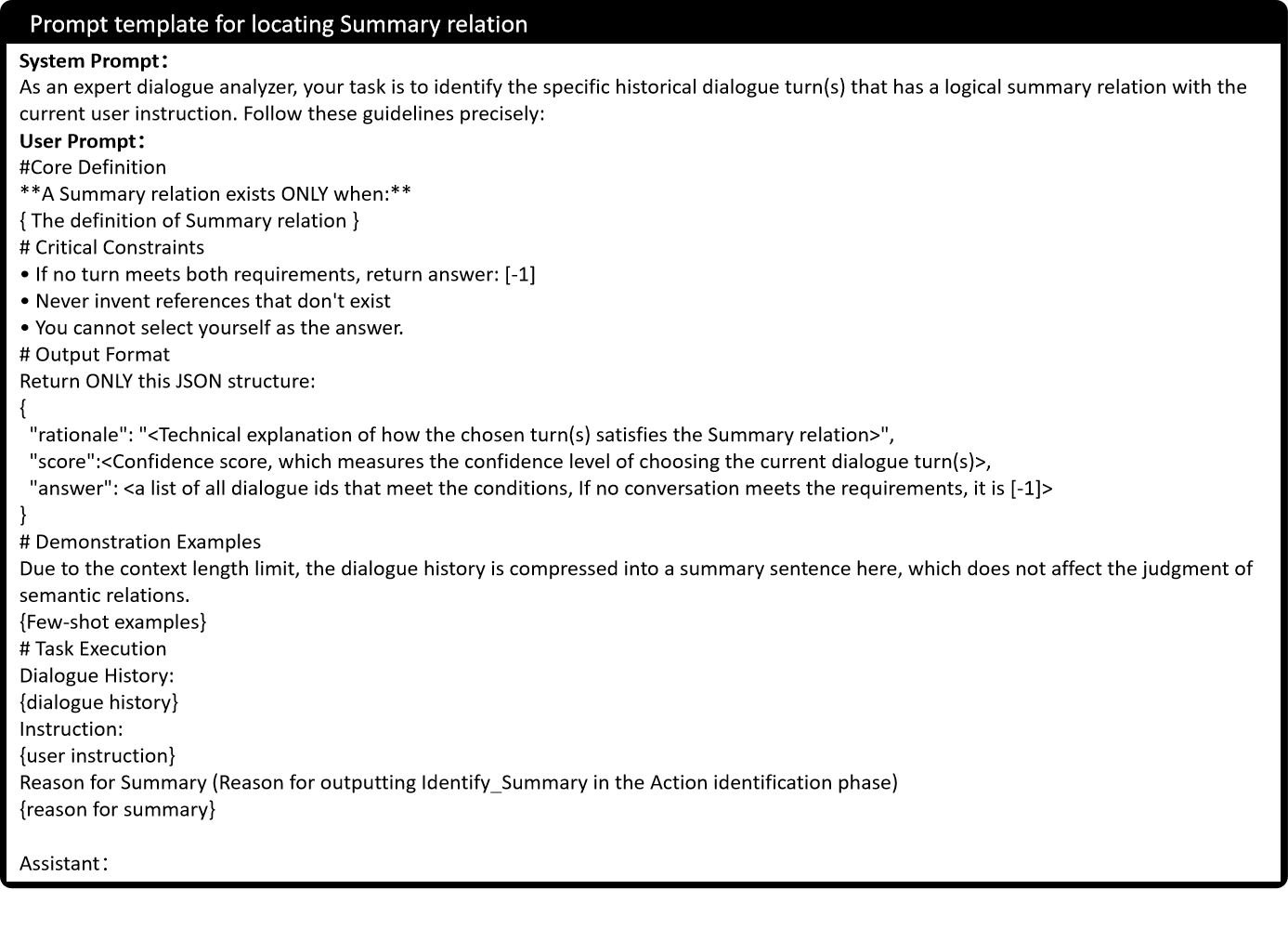}
    \caption{Prompt template for locating Summary relation.}
    \label{fig:identify_summary}
\end{figure}

\begin{figure}[htbp]
    \centering
    \includegraphics[width=1\linewidth]{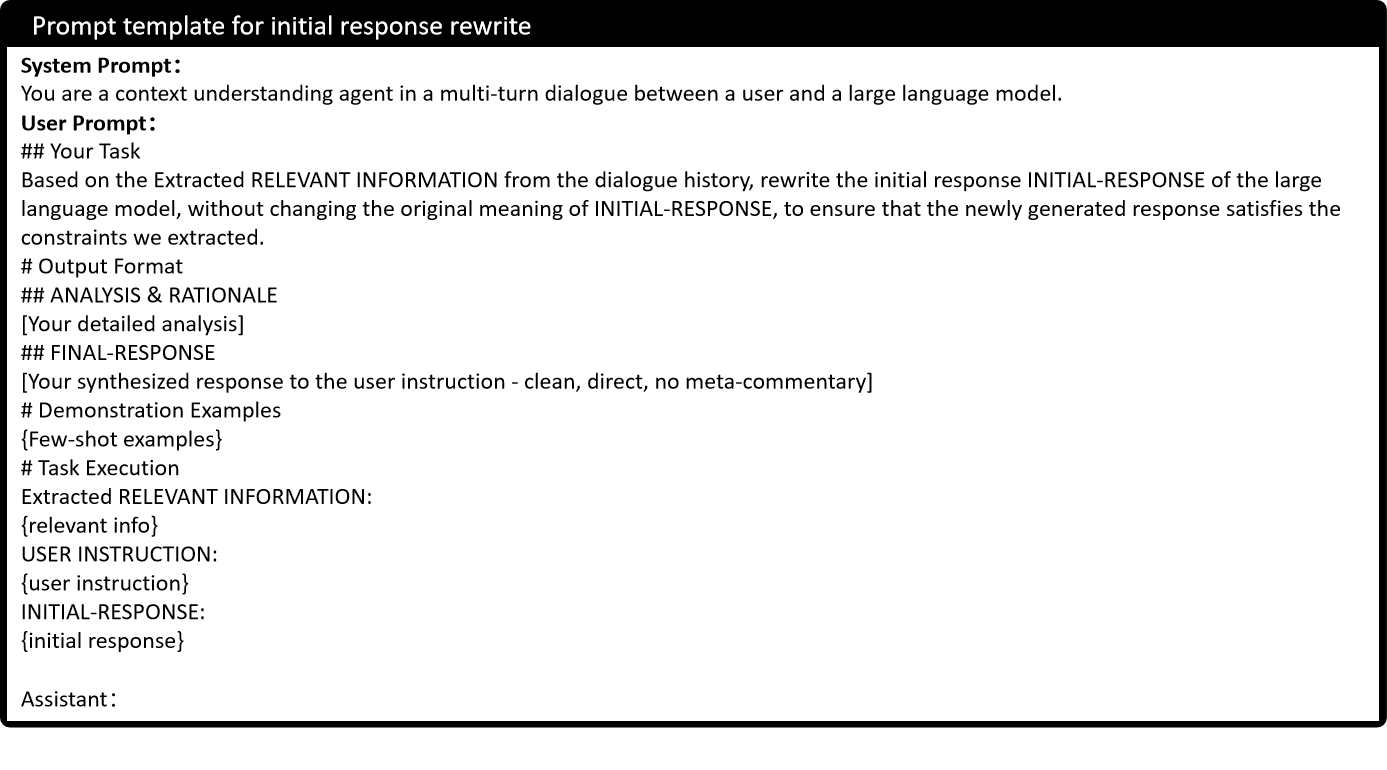}
    \caption{Prompt template for initial response rewrite.}
    \label{fig:response_rewrite}
\end{figure}

Figure~\ref{fig:action_identification} shows the prompt template for Action Identification. Figure~\ref{fig:identify_modify} shows the prompt template for locating the Modify relation(The same applies to Context-Anchored relation). Figure~\ref{fig:choose_topic} shows the prompt template for choosing the topic.
Figure~\ref{fig:identify_summary} shows the prompt template for locating the Summary relation.
Figure~\ref{fig:response_rewrite} shows the prompt template for initial response rewrite.

\end{document}